\documentclass{article}
\usepackage{iclr2021_conference,times}

\usepackage{hyperref}
\usepackage{url}

\iclrfinalcopy

\hypersetup{colorlinks=true,citecolor=darkblue, linkcolor=darkblue, urlcolor=darkblue}

\usepackage{amsmath}
\usepackage{tabularx}
\usepackage{graphics}
\usepackage{lipsum}
\usepackage{multirow}
\usepackage{caption}
\usepackage{colortbl}
\usepackage{amssymb}
\usepackage{float}
\usepackage{subcaption}
\usepackage{wrapfig}
\usepackage{booktabs}
\usepackage{longtable}
\usepackage{amsfonts}
\usepackage{dsfont}
\usepackage{appendix}
\usepackage{xspace}
\usepackage{natbib}

\newcommand{\sfd}[0]{\textsc{Simple facts}\xspace}
\newcommand{\repository}[0]{\url{https://github.com/facebookresearch/multisense_consistency}\xspace}
\usepackage{adjustbox}

\usepackage[noabbrev,capitalize,nameinlink]{cleveref}
\crefname{section}{Section}{Sections}
\crefname{table}{Table}{}
\crefname{figure}{Figure}{}
\crefname{section}{\S}{\S\S}
\Crefname{section}{\S}{\S\S}

\definecolor{darkblue}{rgb}{0, 0, 0.5}

\usepackage{titlesec}
\titleformat{\paragraph}[runin]{\bf\normalsize}{\theparagraph}{1em}{}[. \mbox{}]

\begin{document}

\title{From Form(s) to Meaning: Probing the Semantic Depths of Language Models Using Multisense Consistency}

\author{Xenia Ohmer, Elia Bruni\thanks{Shared senior authorship.}\\
Osnabrueck University\\
\texttt{\{xenia.ohmer,elia.bruni\}}\\
\texttt{@uni-osnabrueck.de}
\And
Dieuwke Hupkes$^*$\\
Meta\\
\texttt{dieuwkehupkes@meta.com}
}

\maketitle


\begin{abstract}
The staggering pace with which the capabilities of large language models (LLMs) are increasing, as measured by a range of commonly used natural language understanding (NLU) benchmarks, raises many questions regarding what ``understanding'' means for a language model and how it compares to human understanding.
This is especially true since many LLMs are exclusively trained on text, casting doubt on whether their stellar benchmark performances are reflective of a true understanding of the problems represented by these benchmarks, or whether LLMs simply excel at uttering textual forms that correlate with what someone who understands the problem would say.
In this philosophically inspired work, we aim to create some separation between form and meaning, with a series of tests that leverage the idea that world understanding should be consistent across presentational modes -- inspired by Fregean \emph{senses} -- of the same meaning.
Specifically, we focus on consistency across languages as well as paraphrases.
Taking GPT-3.5 as our object of study, we evaluate multisense consistency across five different languages and various tasks.
We start the evaluation in a controlled setting, asking the model for simple facts, and then proceed with an evaluation on four popular NLU benchmarks.
We find that the model's multisense consistency is lacking and run several follow-up analyses to verify that this lack of consistency is due to a sense-dependent task understanding.
We conclude that, in this aspect, the understanding of LLMs is still quite far from being consistent and human-like, and deliberate on how this impacts their utility in the context of learning about human language and understanding.
\end{abstract}



\section{Introduction}\label{sec:introduction}

In the past ten years, the abilities of neural language models (LMs) have developed at a -- for most -- unimaginable pace.
This progress has aroused much excitement amongst both scientists and applied researchers, and it comes with a range of interesting questions in various domains.
One category of such questions pertains to the type of (linguistic) intelligence that neural networks possess and how studying them may help us make progress on scientific questions related to linguistics, cognitive science, and human language processing~\citep[e.g.][]{baroni2023role,linzen_baroni_2021,hupkes2020hierarchy,pavlick_2023}.
Specifically recurrent neural networks~\citep{elman1990finding}, which were originally proposed as alternative theories of human sequential processing, have been examined in this context, primarily with respect to topics in syntax and morphology~\citep[among many others][]{dankers-etal-2021-generalising,lakretz2021cognition,jumelet-etal-2021-language,malouf2017abstractive,van2018modeling,abnar-etal-2019-blackbox}.
More recently, their attention-based counterparts have also gained popularity in exploring human linguistic processing~\citep[e.g.][]{timkey2023language,lakretz-etal-2022-transformers}.
In the fields of cognitive science and psychology, neural networks have, among other things, taken on an important role in the debate about syntactic nativism.
In particular later generations of neural networks, which show strong command of natural language syntax~\citep[for an overview, see][Section 3]{chang2023language}, are by some considered to provide a counter-argument to the claim that innate biases are required to learn natural languages~\citep[][i.a.]{Contreras_Kallens2023-uu, piantadosi_2023_LLMs_chomsky, mahowald2023dissociating}.

While the debate on this has hardly been resolved\footnote{
A frequently mentioned critique of this ability is that LMs require vastly more data than humans to arrive at this level of performance (see, e.g. \citet{DUPOUX201843}, or \citet{warstadt2022artificial} for a discussion).
Therefore, more and more research is being carried out to study which syntactic skills language models can learn from smaller amounts of data~\citep{zhang-etal-2021-need}, or even amounts comparable to what children have ingested~\citep{babylm_challenge}.
}
-- and likely will not be for a long time -- LMs have arrived at a stage where their mastery of syntax is almost undisputed, as they obtain nearly perfect scores on syntactic datasets that are challenging even for humans~\citep{wang-2019-etal-superglue,kocijan2023defeat,liang2023holistic}.
In recent times, research exploring the capabilities of (large) language models -- (L)LMs -- has therefore shifted to their ability to correctly process semantics.
In this vain, many datasets have been developed to quantify the extent to which LMs are able to conduct a range of different ``natural language understanding'' (NLU) tasks~\citep[e.g.][]{wang-etal-2018-glue,wang-2019-etal-superglue,hendrycks2021measuring}.
In the literature, there is considerable discussion about the extent to which these datasets accurately measure what they claim to measure.
Commonly used arguments centre around the concept of construct validity, and are supported by findings that datasets contain biases~\citep{gururangan-etal-2018-annotation,benchekroun2023worldsense}, can be solved with heuristics rather than understanding~\citep{mccoy-etal-2019-right,saxon-etal-2023-peco,sen-saffari-2020-models,niven-kao-2019-probing} or do not agree with other datasets claiming to measure the same skill~\citep{sun-2023-validity}.
A much less frequently discussed topic is what this new wave of models, which according to many learn under vastly different circumstances than humans, can still teach us about human language (processing).

While training on inconceivable amounts of data likely makes modern LLMs less suitable to study questions related to syntactic processing and grammar, their new-found natural language understanding abilities open the door to studying a new realm of questions, related to the nature of meaning and how language expresses it.
Some have argued that it is \emph{a priori} not possible to learn meaning from form alone~\citep[e.g.][]{bender-koller-2020-climbing}, yet others disagree or argue that the training signal for at least some LLMs goes beyond form~\citep[e.g.][]{piantadosi2022meaning,mollo2023vector,pavlick_2023,mandelkern2023language}.
Here, we take a different stance: although our approach is embedded in theoretical arguments about the concept of meaning, we propose an empirical method to investigate the notion of meaning acquired when (mostly) being exposed to form.
Our focus is not on explaining \textit{how} meaning is acquired from form, but rather on individuating necessary criteria for grasping meaning and developing a metric to quantify this in LLMs.

Our method is inspired by the seminal works of \citet{frege_1892} and \citet{wittgenstein2010philosophical}, who both put forward influential philosophical theories of meaning.
Frege's work starts from the observation that if the meaning of a word or phrase  were uniquely determined by what it denotes, this would imply that the statements ``a=a'' and ``a=b'' were equally informative, which is evidently not the case, even if a and b refer to the same object.
To solve this apparent paradox, Frege introduced the key concept of the ``sense'' (Sinn) of an expression, which conveys the mode of presentation by which a particular phrase denotes a referent.
As such, Frege's work acknowledges and formalises the idea that different linguistic expressions can share the same referent.
We combine Frege's notion of sense with Wittgenstein's idea that the meaning of language is defined by the effect it has on the world~\citep{wittgenstein2010philosophical}, which thus functions as an anchor for diverse linguistic forms.
Put together, this suggests that having a genuine understanding of language entails understanding its relation to the world, which would in its turn imply consistency among different linguistic expressions that pertain to the same entities within the world.
As LLMs are trained without direct access to the anchor that is the world, we propose that their understanding can be tested by investigating if they -- nevertheless -- have constructed their representational space such that they respond consistently across different forms with the same meaning.

We translate this idea into a method to probe the semantic depths of the form-driven meaning inquired by LLMs, which we call multisense consistency.\footnote{It is worth pointing out that, according to Frege, different linguistic expressions with the same referent may also have the \emph{same} sense. Our borrowing of the term is, in that sense, loose.}
Crucially, we do not presuppose that particular linguistic expressions have the same meaning, but we ask the model itself to generate meaning-preserving expressions, thus focusing more on whether a model has acquired a notion of meaning than on whether that notion is exactly aligned with ours.
If a model generates consistent responses when prompted with these expressions, this would suggest it might be linking them to their common underlying meaning.
We apply our consistency-based test to investigate one of the currently most advanced models: GPT-3.5.\footnote{Note that GPT-3.5 was trained on more than form. While the details are unknown, the training involved Reinforcement Learning from Human Feedback~\citep{rlhf}, which arguably provides additional information such as communicative intent. It has also been argued that, even without this additional training stage, typical training corpora contain information beyond form, for example, written computer programs and the outputs they generate~\citep{bender-koller-2020-climbing}. Detecting inconsistencies thus suggests that even this kind of additional information does not give rise to a meaning-based understanding. Beyond that, multimodal LLMs, which we do not consider here, encounter more explicit information about form-meaning mappings during training.}
In a series of experiments, beginning with the evaluation of basic truth-conditional statements and progressing to more complex ones, we discover numerous instances where the LLM responds inconsistently across different, meaning-preserving expressions, even in scenarios as straightforward as reiterating a fact.
This is true both when meaning-preserving senses are \emph{paraphrases} and \emph{translations}.
Our results, which we substantiate with several follow-up analyses, illustrate that even one of the best performing LLMs does not seem to have meaning-preserving representations that align with what a Fregean theory of meaning may consider true meaning.
While this may come as no surprise to many, it still begs the question of what the conclusion would have been if the model \emph{did} pass this consistency-based test, and if there is anything that could convince us that an LLM has -- in fact -- truly acquired meaning.
We elaborate on this in our discussion.

\paragraph{Outline}

In the remainder of this paper, we will first take a closer look at Frege's theory on sense and reference, which provides the framework for our approach (\cref{sec:philosophical_background}).
We will then give a high-level overview of how multisense consistency can be used to study the discrepancy between competence in form and competence in meaning (\cref{sec:procedure}) before providing more details on our experiments, such as the model and the senses considered (\cref{sec:experimental_details}).
We discuss results for two different types of datasets -- simple hand-crafted probes of factual knowledge and popular NLU benchmarks (\cref{sec:multilingual_factual_consistency} and \cref{sec:benchmark_results}, respectively), following up with several analyses to study when and why inconsistencies arise (\cref{sec:analyses}).
Finally, we position our contribution in the context of related work (\cref{sec:related_work}) and discuss our findings within the broader scope of using LLMs as models of meaning (\cref{sec:discussion}).


\section{Philosophical background}\label{sec:philosophical_background}

Our study draws inspiration from philosophical notions of meaning, in particular the one put forth by \citet{frege_1892}.
Here, we provide a short discussion of this philosophical backbone and its relevance to evaluating LLMs.

\begin{figure}[t]
    \includegraphics[width=\textwidth]{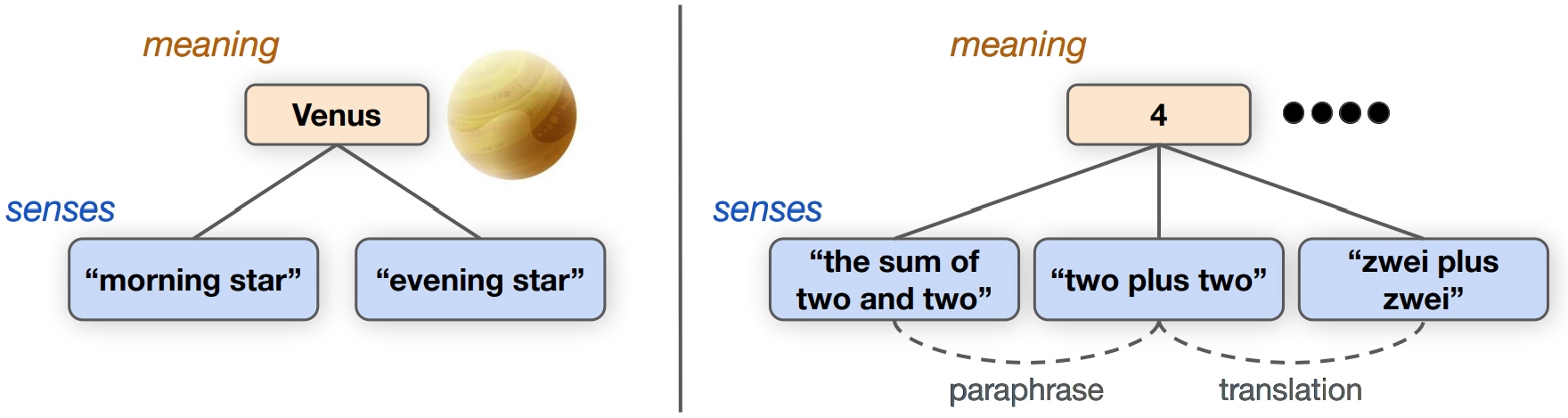}
    \caption{\textbf{Illustration of the relationship between sense and meaning for the classical Fregean example of ``morning star'' and ``evening star'' (left) and for the \textsc{addition} task in our experiments (right).}\protect\footnotemark}\label{fig:sense_meaning}
\end{figure}

\paragraph{Sense and reference}
Before Frege, theories of meaning often struggled to explain the relationship between words and the world they describe, typically approaching this relationship in a linear and simplistic way.
These theories faced difficulties in explaining how language could meaningfully refer to non-existent entities, define the meaning of statements that cannot be easily mapped to a truth value, or handle identity statements where two different expressions appear to refer to the same object.
Frege's introduction of the concepts of ``sense'' (Sinn) and ``reference'' (Bedeutung) offered a solution to these problems.
The reference of an expression is the actual entity or concept the expression corresponds to in the real world and is decisive in determining the truth value of a sentence.
The sense of an expression, in contrast, comprises the way in which this reference is presented.
For example, the morning star and the evening star refer to the same celestial body, Venus, but have different senses (see \cref{fig:sense_meaning}).
Not only can the same reference be presented through different senses, the same sense can also be realised through different expressions -- with some surface level variations.
If two forms (expressions) have the same sense, it is possible to determine a priori that they map to the same referent.
However, if two forms have different senses, learning that they have the same referent provides an extension of our knowledge.
The distinction between sense and reference is vital for understanding identity statements and language paradoxes, where the same reference may be approached through distinct senses.
Furthermore, it implies that language is not just a tool for naming or describing things but serves as a window into how speakers conceptualise and engage with their environment.
By distinguishing between sense and reference, Frege provided a framework that could handle the subtleties of language use, such as ambiguity, metaphor, and the context-dependent nature of meaning.
This framework, now central to the philosophy of language, underscores that a certain reference can be expressed and conceptualised in different ways.

\footnotetext{The illustration of Venus was taken from \url{https://www.universiteitleiden.nl/leven-in-het-heelal/over-leven/venus}.}

\paragraph{Relevance to LLMs}
Making use of the conceptual groundwork laid by Frege, we posit that true linguistic understanding in LLMs should be evident not just in processing the surface form of text but in grasping the reference that underlies this text.
Our methodology leverages this principle by examining the model's consistency across different expressions that refer to the same underlying meaning.
By using the model itself to generate the alternative forms, we ensure that it should -- in principle -- ``know'' that they have the same meaning.
Taking the example above, if a person is not aware that ``evening star'' and ``morning star'' have the same reference (or ``two plus two'' and ``the sum of two and two'' for that matter), their response to these two expressions will likely not be the same.
However, if a person knows that the two expressions can be used interchangeably, they should be able to answer the same facts about Venus regardless of the choice of expression.
By testing across languages and paraphrases, we essentially probe whether LLMs can discern that different textual forms (or senses) may converge on the same reference or meaning, thus revealing a more profound understanding of language beyond mere textual mimicry.

Adopting a loose interpretation of Frege's notion of ``sense'', our multisense consistency method applies to the more general case of different senses as well as the more specific case of different forms expressing the same sense.
At the same time, considering translations and paraphrases as potentially involving shifts from one sense to another acknowledges the complexity and richness of language.
Different languages and (paraphrased) expressions can present the same referent (or truth value) in diverse ways, capturing the many-sided nature of human thought and culture.
Regardless of shifts in sense, the crucial factor is the preservation of the reference -- the actual object or truth condition the expressions pertain to.
This approach is consistent with Frege's emphasis on the importance of reference in determining the truth value of sentences.



\section{Evaluating multisense consistency}\label{sec:procedure}

\begin{figure}[]
    \centering
    \includegraphics[width=\textwidth]{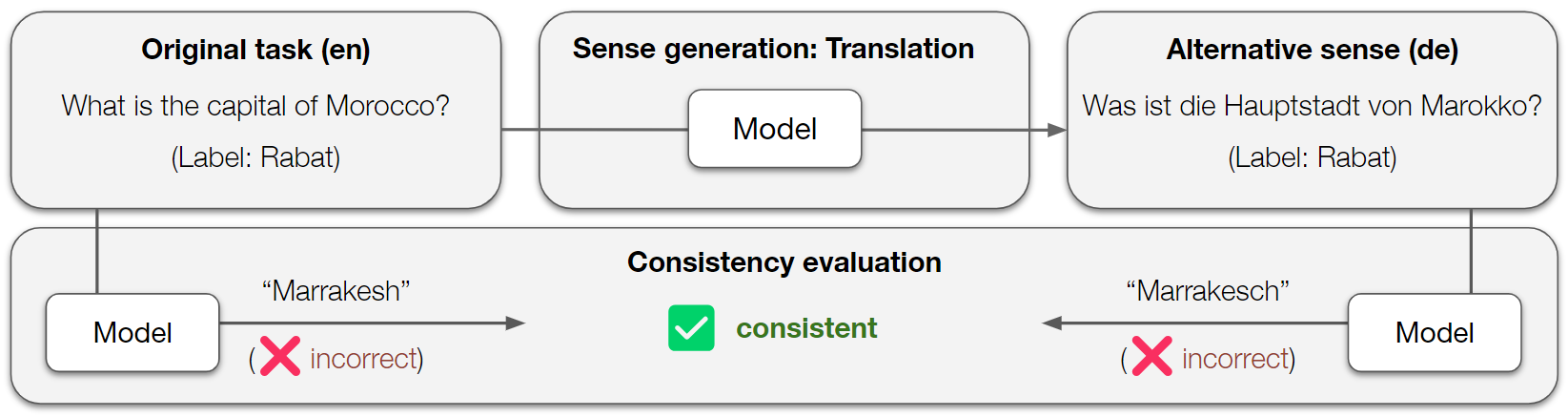}
    \caption{\textbf{Illustration of the multisense consistency paradigm}.
    We use a model to generate alternative meaning-preserving senses of the original input, and then evaluate whether the same model gives consistent responses to the original input and alternative sense.
    In this example, the task is to answer a simple factual question, and the model is asked to generate an alternative sense through translation (from English to German).
    The example illustrates that accuracy and consistency are distinct.
    Even though the model's responses are incorrect (Marrakesh/Marrakesch instead of Rabat), they are consistent because they refer to the same city.}
    \label{fig:paradigm}
\end{figure}

Concretely speaking, we investigate whether LLMs can be considered to have a form-independent notion of meaning by constructing a test that quantifies whether their understanding is consistent across different expressions with the same meaning.
In what follows, we refer to those tuples of expressions as \emph{senses}.
Before diving into our experiments, we first give a high-level overview of the main components of this idea.
We discuss how we generate different senses (\cref{subsec:methods-generating-senses}), what data we start from to do so (\cref{subsec:methods-base-data}), and our method for computing multisense consistency (\cref{subsec:methods-measuring-self-consistency}).
We provide a schematic in \cref{fig:paradigm}.

\subsection{Generating different senses}\label{subsec:methods-generating-senses}

The first important component of our paradigm are senses: tuples of expressions that express the same meaning in different manners.
Senses could be generated in several ways.
In this work, we consider two different methods: translation and paraphrasing, which we will denote by the superscripts $^T$ and $^P$, respectively.
Importantly, we use the model under investigation to generate meaning-preserving senses, with the idea that \emph{if} the model has a meaning-based understanding and is proficient at generating alternative senses (which we control for in \cref{sec:analyses}), these senses should have the same meaning according to the model and should thus elicit consistent responses.
On the contrary, if a model's meaning is tied to a specific form, there is no reason to assume the response to two senses that have the same meaning should be the same.
Thus, using the model to generate the senses controls for subjective meaning-consistency.
This approach mirrors Frege's seminal distinction between sense and reference~\citep{frege_1892} emphasising that true understanding transcends linguistic form to grasp the underlying meaning.
Just as Frege illustrated how different expressions can denote the same reference, our paradigm tests whether LLMs can discern and maintain this crucial distinction in a computational context.

\subsection{The ``base'' data}\label{subsec:methods-base-data}
The second component of our paradigm is a ``base'' dataset, to generate different senses from.
While the multisense consistency paradigm can in theory be applied to any data, generating senses that have the same meaning may be more or less difficult depending on the initial data and the sense-generation procedure.
In this paper, we work with two types of datasets.
The first type are synthetically constructed datasets with simple facts.
Because we can be certain that their meanings are consistent across languages, they allow us to test form-independent meaning in a very controlled way.
We describe this data as well as our experiments with this data in \cref{sec:multilingual_factual_consistency}.
Secondly, we consider benchmarks commonly used to evaluate understanding in LLMs.
Specifically, we include four different benchmarks covering four different types of NLU tasks: PAWS~\citep{zhang-etal-2019-paws} for paraphrase detection, the English portion of XNLI~\citep{conneau-etal-2018-xnli} for natural language inference, COPA~\citep{roemmele2011choice} for commonsense (causal) reasoning, and Belebele~\citep{bandarkar2023belebele} for reading comprehension.
We describe this data as well as our experiments with this data in \cref{sec:benchmark_results}.

\subsection{Measuring self-consistency}\label{subsec:methods-measuring-self-consistency}

Lastly, given two senses with the same meaning and two model responses to those senses, we need to define when those two responses are considered to be the same.
In other words, we need to specify a method to compute consistency.
Consistency is distinct from accuracy or other performance metrics, in that the model's responses to one sense are evaluated against its responses to the other sense, rather than the ground truth (see~\cref{fig:paradigm}).
Whether responses count as consistent depends both on the task and the way that different senses are generated.
For instance, if senses are generated through paraphrasing and the task is a classification task where a model has to pick an answer from a predefined list (e.g.\ ``yes''/``no''), exact match (EM) is a good candidate to quantify consistency.
If senses are generated through translation, however, model answers will likely be given in different languages, and may look completely different but still share a meaning (e.g. ``yes'' in English, ``ja'' in German).
In that case, a more custom consistency function is required to judge consistency across senses.
For open-ended generation tasks, it can be complicated to define consistency.
In such cases, an option can be to ask the model itself to judge whether its two answers have the same meaning.
In our experiments, we use different methods to evaluate consistency, which we elaborate upon in the respective sections.

\subsection{Summary of overall procedure}\label{subsec:overall-procedure}

Overall, our procedure can be summarised as follows.
Given a model $\mathcal{M}$ and a task $\mathcal{T}$, which consists of datapoints $\mathcal{T} = \lbrace x_1, \dots, x_n \rbrace$,
\begin{enumerate}
    \item Collect the model's responses on $\mathcal{T}$: $\:R = ( r_1, \dots, r_n )$, with $r_i = \mathcal{M}(x_i)$.
    \item Use the model to generate an alternative sense $\mathcal{T}^*$ of the task, using a specific prompt $p$: $\:\mathcal{T}^* = \lbrace x_1^*, \dots, x_n^* \rbrace$, with $x_i^* = \mathcal{M}(p, x_i)$.
    \item Collect the model's responses on $\mathcal{T}^*$: $\:R^* = ( r_1^*, \dots, r_n^* )$, with $r_i^* = \mathcal{M}(x_i^*)$.
    \item Calculate the consistency between $R$ and $R^*$ according to some function: $C(R, R^*) = \frac{1}{n} \sum_{i=1}^n f(r_i, r_i^*)$.
\end{enumerate}
\noindent The resulting consistency value $C$ expresses multisense consistency.



\section{Experimental details}\label{sec:experimental_details}

Before getting to our experiments, we provide some basic details about the setup that all experiments share.

\subsection{Model}\label{subsec:experiments_models}
We investigate \textsc{gpt-3.5-turbo-0613}, a specific snapshot of \textsc{gpt-3.5-turbo} from June 13th, 2023.
We use the default parameters but set the temperature to $0.2$.
The sampling temperature can be chosen between $0$--$2$, and $0.2$ is considered a low value, leading to more deterministic and focused output (see also the OpenAI API documentation\footnote{\url{https://platform.openai.com/docs/api-reference/chat/create}}).
In our case, a small temperature yields model responses that closely match the template answers for benchmarking, as well as model translations that closely preserve the meaning of the source sentences.

\subsection{Senses}\label{subsec:experiments_languages}

In all our experiments, our starting point is an English dataset, which we denote with $en$.
We consider model-generated paraphrases of that data and model-generated translations into other languages.
For some datasets, we also have external translations, which we use for saliency checks and comparisons.
Target languages include German ($de$), Italian ($it$), Dutch ($nl$), and Swedish ($sv$).
We use the current common crawl statistics\footnote{\url{https://commoncrawl.github.io/cc-crawl-statistics/plots/languages}, CC-MAIN-2023-40} to compute an estimate of how low- or high-resource these languages are in web-based corpora.
Of this corpus, English constitutes 46\% of the data, German 5.8\%, Italian 2.7\%, Dutch 2.2\%, and Swedish 0.7\%.
We assume that the GPT-3.5 training data qualitatively follows a similar pattern for these languages, from higher- to lower-resource.
The multisense evaluation method only works if the model is able to accurately paraphrase and translate the inputs.
Therefore, we do not include even lower-resource languages.
With our selection of languages, we aim to cover some range in the amount of training data without compromising translation quality.

\subsection{Same-sense baseline}\label{subsec:experiments_baseline}
We report multisense consistency next to a same-sense baseline consistency.
The baseline consistency is the consistency between two generations with the exact same English input ($id$).
In other words, the two inputs underlying the baseline consistency do not even differ in form.
Differences in model responses on these inputs can thus be attributed to inherent model stochasticity (possible because of the non-zero sampling temperature).
The baseline consistency therefore serves as a reference, which can be used to estimate the degree to which inconsistencies between different senses can be attributed to differences in form rather than such inherent stochasticity.




\section{Multilingual factual consistency}\label{sec:multilingual_factual_consistency}

In our first set of experiments, we test the model's form-dependency when answering simple questions about facts.
To do so, we generate datasets that assess a model's consistency in representing basic factual information of various knowledge domains.
The power of these datasets lies in their simplicity.
There is little room for nuances in wording across different senses that could cause the model to assign a different meaning.
Factual knowledge -- in contrast to more complex aspects such as expressions of sentiment -- is easy to keep stable across senses, because the meaning of factual statements collapses to their truth value. 
To give an example, if you ask a colleague who is fluent in both French and English if a particular statement is true, you expect their answer to be invariant to the language (French or English) in which you ask this question.
Along the same lines, the model should generate consistent responses when asked about the kinds of simple facts considered here.
Given that the fact-based questions leave hardly any room for ambiguity, inconsistent responses point straight to a form-dependent ``understanding''.

\subsection{Methods}\label{subsec:simple_facts_data}

Our \sfd dataset consists of five distinct datasets, each containing one or more subtasks.

\paragraph{Dataset creation}
\cref{tab:templates} provides an overview of the datasets and subtasks, including information on the dataset size and examples.
Each dataset comprises a single template with specific content fields masked out.
During dataset creation, different entities (names, dates, etc.) are inserted into these fields.
For instance, the \textsc{writers} dataset is based on the template ``In what year was the writer [WRITER] born?'' and in each datapoint, [WRITER] is replaced by the name of a different writer.
For both \textsc{writers} and \textsc{companies}, we ensure -- with some simplification -- that the writers and companies are evenly distributed over countries in which the languages we consider constitute the dominant language.\footnote{The languages we consider are spoken in different countries, but we tend to focus on one country each. 
For example, for \textsc{companies}, we consider an equal amount of US, German, Italian, Dutch, and Swedish companies, establishing a rough correspondence between prompt languages and factual information.}
More details on each dataset can be found in Appendix~\ref{app:simple_facts}.

\paragraph{Sense generation}
We prompt the model to generate different senses for each (sub)task by asking it to paraphrase or translate the corresponding template.
Because only the template changes, we can evaluate the quality of the generated paraphrases and translations by hand.
Details on the instructions used for generating different senses can be found in Appendix~\ref{app:B_sense_creation} and the original instructions and the model's translations can be found in Appendix~\ref{app:C_instructions}.

\paragraph{Model instructions} To facilitate the performance and consistency evaluations, we always instruct the model to respond with a single entity (e.g.\ the name of the athlete for \textsc{olympics}) or number (e.g.\ ``4754'' for \textsc{arithmetics}).%
\footnote{We double-checked if the model sometimes indicates that it does not know the correct answer, if it is not instructed to respond in these particular ways.
On all datasets but \textsc{writers}, it does so very rarely ($\le$ 1\%).
Besides, a comparison of the model's responses to \textsc{writers} in $en$ and $de^T$ showed that even if the model indicates that it does not know the correct answer, it does not do so consistently between senses.}
On the \textsc{arithmetics} dataset, the model is further instructed to reply with the numerical answer, even though the two summands are spelled out.

\begin{table}[]
\caption{\textbf{Simple facts datasets.} In this table, we provide the templates we used to generate the \textsc{simple facts} datasets, and the total number of examples in each dataset (N). For each template, we provide an example in which the mask(s) are populated with an example datapoint (in bold) from our datasets.}\label{tab:templates}
\centering
\begin{adjustbox}{max width=\textwidth}
\begin{tabular}{l|l|l|l}
\toprule
\textbf{dataset} & \textbf{subtask} & $\mathbf{N}$ & \textbf{template / example} \\
\midrule
\small\textsc{arithmetics}                  & -          & $500$  &   \small ``What is \textbf{three hundred seventy-five} plus \textbf{twenty-three}?''\\
\midrule
\multirow{3}{*}{\small\textsc{elements}}    &  \small \textsc{from-element}     & $90$  &   \small ``What is the atomic number of the chemical element \textbf{He}?'' \\
\cline{2-4}
                                            &   \multirow{2}{*}{\small\textsc{from-position}}   & \multirow{2}{*}{$90$}  &   \small ``What is the atomic number of the chemical element \\
                                            &                                   &       &   \small  in period \textbf{5} and group \textbf{7}?'' \\
\midrule
\multirow{2}{*}{\small\textsc{olympics}}    &   \multirow{2}{*}{\small\textsc{100m}}            & \multirow{2}{*}{$148$}   &   \small ``Who won the \textbf{gold} medal in the \textbf{men's} 100 meters at the \\
                                            &                                   &        &   \small  \textbf{2000} Summer Olympics?'' \\
\cline{2-4}
                                            &   \multirow{2}{*}{\small\textsc{downhill}}        & \multirow{2}{*}{$117$}   &   \small ``Who won the \textbf{bronze} medal in the \textbf{women's} downhill \\
                                            &                                   &        &   \small   competition at the \textbf{1976} Winter Olympics?'' \\
\midrule
\small\textsc{writers}                      &  -                                & $186\times 5 = 930$  & \small ``In what year was the writer \textbf{Friedrich Schiller} born?'' \\
\midrule
\small\textsc{companies}                    &  -                                & $100\times 5 = 500$  & \small ``In what city does \textbf{Airbus SE} have its headquarters?'' \\
\bottomrule
\end{tabular}
\end{adjustbox}
\end{table}

\paragraph{Consistency evaluation}
The \textsc{simple facts} datasets contain a set of correct answers $A_d$ for each datapoint $d$.
For example, the answer sets for \textsc{companies} cover all variations in city names for the languages we work with (e.g., for the city of Berlin $A_d = \lbrace$ ``Berlin'', ``Berlijn'', ``Berlino''$\rbrace$).
To give another example, the answer sets for \textsc{olympics} contain different variations of the athletes' names (e.g., for the winner of the men's hundred metres in 1920 $A_d = \lbrace$``Charlie Paddock'', ``Charles Paddock'', ``Charles William Paddock''$\rbrace$) as well as multiple names if there is more than one winner.
Model responses are always normalised by lowercasing and removing surrounding white spaces and punctuation.
Given the normalised models responses, $R$ and $R^*$, the consistency
\begin{equation*}
    C(R, R^*) = \frac{1}{n} \sum_{i=1}^n f(r_i, r_i^*)\:\label{eq:consistency1}
\end{equation*}
(see \cref{subsec:overall-procedure}, step 4) is calculated as
\[
    f(r, r^*) =
\begin{cases}
    \mathds{1}_{r \in A \:\&\: r^* \in A},& \text{if } \exists A:r\in A$ or $r^*\in A \:,\\
    \mathds{1}_{r = r^*},             & \text{else}\:;
\end{cases}
\]
where $A$ is a set of possible answers for a datapoint in $\mathcal{T}$ (which are the same as the answer sets for $\mathcal{T}^*$) and $\mathds{1}$ is the indicator function.
In other words, if an answer set is available that contains the model's response $r$ or $r^*$, both of the responses have to be in that set to be consistent.
If no such set exists, consistency is approximated by exact match.

\subsection{Results}\label{subsec:simple_facts_results}
Before studying the model's consistency, we consider its ability to correctly answer the factual questions.
The model's performance helps us put its consistency into perspective because it sets an upper and a lower bound for the consistency.
For instance, if a model reaches maximal performance across senses on some task, it will also be perfectly consistent.

\begin{figure}[]
    \centering
    \includegraphics[width=0.9\textwidth]{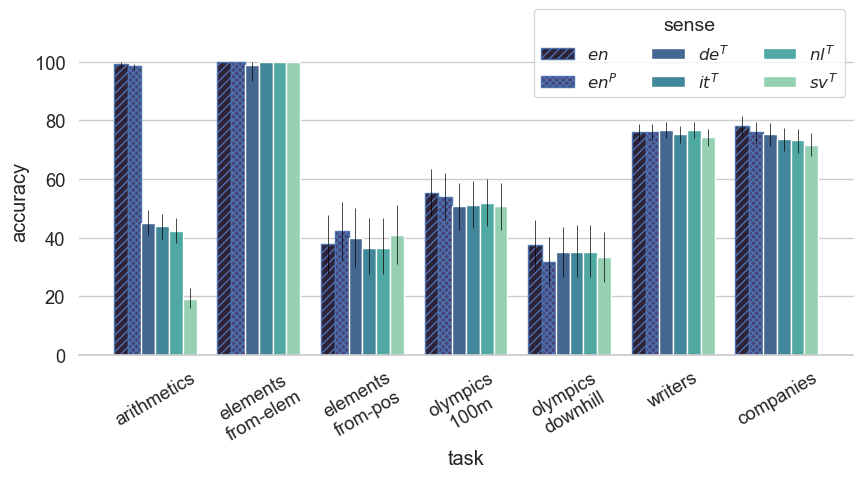}
    \caption{\textbf{Accuracy (\%) for the \sfd datasets, with 95\% confidence intervals.} Apart from the arithmetics task, the accuracy scores are generally similar across different senses. Numerical scores can be found in \cref{tab:simple-facts-accuracy}.}
    \label{fig:simple-facts-accuracy}
\end{figure}

\paragraph{Performance}
We compute the accuracy (exact match) scores across datasets and senses.\footnote{The model is instructed to reply with the correct entitity and no additional words. In the large majority of the cases the model follows this instruction, such that there is little difference between counting responses as correct when they \emph{contain} the right answer instead of being an \emph{exact match}. For details, see Appendix~\ref{app:containment_versus_exact_match}.}
For some datapoints there are several correct answers; the model's response counts as correct if it corresponds to one of them.
The set of correct answers contains variations in naming (e.g.~``Charles Paddock'', ``Charlie Paddock'', ``Charles William Paddock''), including variations between the languages we use (e.g.~``Berlin'', ``Berlino'', ``Berlijn'').
The full list of equivalent answers can be found in our repository.\footnote{\repository}
In \cref{fig:simple-facts-accuracy}, we can see that the difficulty of the tasks and subtasks varies strongly.
For instance, accuracies on \textsc{elements-from-element} are uniformly close to 100\% whereas accuracies on \textsc{olympics-downhill} are below 38\%.
However, the model's performance within subtasks is relatively consistent across the different senses, except for \textsc{arithmetics}, where performance in English is vastly higher than performance for other languages.

The differences in accuracy for \textsc{arithmetics} are striking.
We double-checked if the model fails to reply with a numerical answer in some of the languages but this was not the case.
In Swedish, the model sometimes responds with the entire equation instead of the correct sum (e.g. ``342 + 122 = 464'' instead of ``464'') but accuracy only increases by 2\% when accounting for these cases.
It could be that spelled-out numbers are rare in the training corpus such that high-versus low-resource effects get magnified, which could explain why there is a big drop from $en$ to $de$/$it$/$nl$, and then another one to $se$.

\begin{figure}[]
    \centering
    \includegraphics[width=0.9\textwidth]{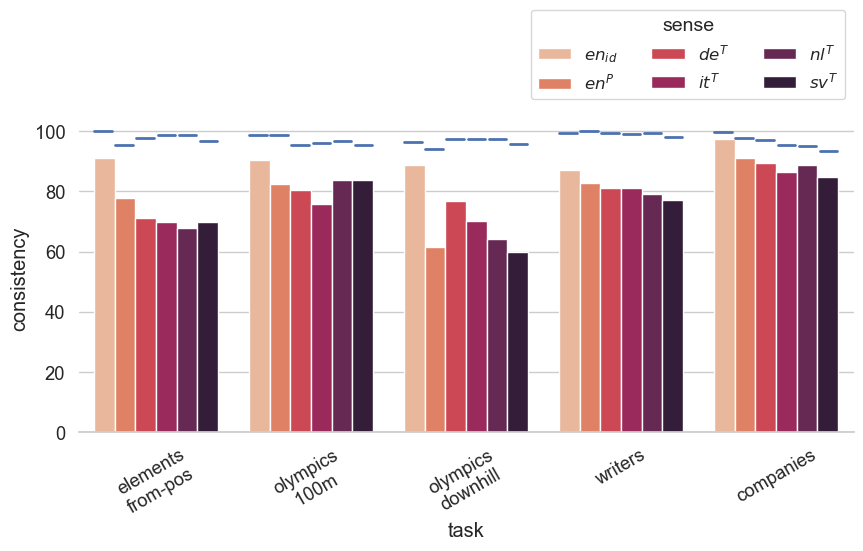}
    \caption{\textbf{Consistency (\%) for the \sfd datasets.}
    None of the senses have a consistency close to the maximum possible given the difference in accuracy between the two senses (indicated by the horizontal blue lines), indicating that the models are inconsistent even beyond those differences. Numerical scores can be found in \cref{tab:simple-facts-consistency}.}
    \label{fig:simple-facts-consistency}
\end{figure}

\paragraph{Consistency}
Next, we consider how consistent the model's representations are across senses.
We report the results in \cref{fig:simple-facts-consistency}.
Because the generation-process is stochastic at non-zero temperature, asking the same question twice may lead to different responses.
We exploit this to report also same-sense consistency between two $en$-runs (denoted with $id$).
Note that if a model has a maximal accuracy on one of the senses, its consistency score equals the accuracy of the other sense, without providing any evidence for form-independent meaning representations. 
We therefore exclude the \textsc{arithmetics} and \textsc{elements-from-elem} task from our consistency results.
More generally, given a difference in accuracy between two senses, $\Delta$(Acc), the consistency cannot be higher than $1-\Delta$(Acc).\footnote{For example, if the model is 80\% correct on one sense and 60\% correct on another sense, the maximal consistency is achieved when the respective overlap between correct and incorrect responses is maximal: The same 60\% of the datapoints are correct on both senses, and the same 20\% of the datapoints are incorrect on both senses, resulting in 100\%-(80\%-60\%)=80\% consistency.}
We indicate these upper bounds in the figure with blue lines above each bar.
While consistency and accuracy are thus not independent, as long as accuracies are not at 100\%, they are clearly distinct.
As we will see, even if the differences between the accuracies are small, the consistency may vary wildly.

In \cref{fig:simple-facts-consistency}, we can see a manifestation of this statement: although the accuracy scores across senses are all comparable (see \cref{fig:simple-facts-accuracy}), there is not a single case where the consistencies are near-maximal. 
This is remarkable given the simplicity of the tasks and instructions.
Even for English paraphrases, consistency can be as low as $61.5$\% at a $88.9$\% baseline (see \textsc{olympics-downhill}).
In this case, almost all inconsistencies arise because the model replies with the names of different athletes, usually winners of other medals in the same competition or winners of other competitions.
For example, when asked for the female bronze medallist in 1988, the model gives the correct answer to the original prompt (``Brigitte Oertli'') but replies with the name of the world champion of 1989 to the paraphrased prompt (``Karin Dedler'').
More examples can be found in Appendix~\ref{appendix:examples}.
The baseline scores ($id$) show that the inconsistencies are not (primarily) caused by the model assigning equal probabilities to possible answers, leading to different outputs on different senses.
While the baseline scores are not maximal, they are much higher than what would be expected in such a case.\footnote{The simple facts datasets are open QA tasks. When the model is asked for an entity (e.g. a city), it can potentially choose its answer from the set of all entities in the relevant category (e.g. all cities). If the model assigned similar probabilities to many answers in this set, it would likely be inconsistent whenever it is incorrect. In that case, the baseline consistency would be less than or at the maximum (when there is a perfect overlap between correct responses) equal to the model's accuracy on $en$.}
In other words, most inconsistencies cannot be attributed to the lack of a clear winner, in which case the model would sample from several roughly equally low probabilities.



\section{Natural language understanding benchmarks}\label{sec:benchmark_results}

Our results with the \textsc{simple facts} datasets point to substantial form-dependencies in the LLM's representation of factual knowledge.
Next, we investigate how the model behaves on a set of different NLU tasks in which meaning and task understanding is more complex than merely reiterating knowledge.

\subsection{Methods}\label{subsec:benchmark_data}

For our continued evaluation of consistency across more complicated scenarios we consider four different benchmarks covering four different types of NLU tasks.

\paragraph{Datasets}
First, we consider PAWS \citep{zhang-etal-2019-paws}, a paraphrase dataset where sentence pairs were adversarially created by word-swapping, resulting in negative pairs that have clearly distinct meanings but high lexical overlap (see, for instance, the example in \cref{tab:benchmark_data}).
Second, we consider (mainly the English portion of) XNLI \citep{conneau-etal-2018-xnli}, a language inference task containing sentence pairs that either entail or contradict each other, or have a neutral relationship.
Third, we use COPA \citep{roemmele2011choice}, a dataset containing tuples of a premise and two alternatives, where the task is to select the alternative that more plausibly has a relation with the premise.
Lastly, Belebele \citep{bandarkar2023belebele} is a reading comprehension task with multiple choice questions where an answer should be given based on a text passage.
We run all our evaluations on the test split of the respective datasets.
Note that all tasks correspond to classification problems; we standardise the model's responses and map them onto the corresponding class labels.
Furthermore, for some of the languages we consider, parallel data for the tasks exists either in the original corpus (in case of Belebele and XNLI) or in multilingual versions of the corpus \citep[PAWS-X and XCOPA,][respectively]{yang-etal-2019-paws,ponti-etal-2020-xcopa}.
While our paradigm does not require parallel multilingual datasets, we use them in \cref{sec:analyses} to run additional analyses.

\paragraph{Sense generation and model instructions} For each dataset, we write an English instruction which together with the task input data forms the prompt presented to the model (see \cref{tab:benchmark_data}). 
We ask the model to paraphrase and translate the instruction and the input data separately, and we recompose the two outputs to generate the alternative sense.
Individual datapoints in the benchmarks comprise several components, for example, a premise and a hypothesis in the case of XNLI.
We provide all these components within the same prompt when asking the model to paraphrase or translate.
Combining the components for each datapoint has the advantage that the resulting paraphrases/translations will be more consistent (e.g., the model will resolve ambiguities or make certain translation choices in the same way across components).
We compared this method to paraphrasing/translating each component separately, and it resulted in slightly higher task accuracies on the generated senses.
More details on the sense generation can be found in Appendix~\ref{app:B_sense_creation}, and the model's translations and paraphrases of the instructions can be found in Appendix~\ref{app:C_instructions}.

\begin{longtable}{@{\extracolsep{5pt}}l|p{11cm}}
    \caption{\textbf{Instructions and example inputs for the benchmark data.} We provide an example for each benchmark dataset in our experiments. The example input is given in bold, the instructions in normal font.}\label{tab:benchmark_data} \\
\toprule
\textbf{dataset}    & \textbf{template / example} \\
\midrule
\small paws        &   \small Do the following two sentences have the same meaning?\newline Sentence 1: ``\textbf{The Tabaci River is a tributary of the River Leurda in Romania .}''\newline Sentence 2: ``\textbf{The Leurda River is a tributary of the River Tabaci in Romania .}''\newline Please reply with a single word, either ``yes'' or ``no''.\\
\midrule
\small xnli (en)   &   \small Given the following premise and hypothesis, please identify whether the premise entails the hypothesis, contradicts the hypothesis, or neither of the two.\newline Premise: ``\textbf{Well, I wasn't even thinking about that, but I was so frustrated, and, I ended up talking to him again.
}''\newline Hypothesis: ``\textbf{I haven't spoken to him again.}''\newline Please reply with a single word: ``entailment'' if the premise entails the hypothesis, ``contradiction'' if the premise contradicts the hypothesis, and ``neutral'' if the premise neither entails nor contradicts the hypothesis.\\
\midrule
\small copa        &   \small Given the following premise, which of the two alternatives is more plausible?\newline Premise: ``\textbf{The item was packaged in bubble wrap.}''\newline Alternative 1: ``\textbf{It was fragile.}''\newline Alternative 2: ``\textbf{It was small.}''\newline Please answer with a single word: ``Alternative-1'' if alternative 1 is more plausible and ``Alternative-2'' if alternative 2 is more plausible.\\
\midrule
\small belebele    &   \small \textbf{Virtually all computers in use today are based on the manipulation of information which is coded in the form of binary numbers. A binary number can have only one of two values, i.e. 0 or 1, and these numbers are referred to as binary digits - or bits, to use computer jargon.}\newline \newline \textbf{According to the passage, which of the following is an example of a five bit binary number?}\newline \newline Option A: \textbf{1010}\newline Option B: \textbf{12001}\newline Option C: \textbf{10010}\newline Option D: \textbf{110101}\newline \newline Please reply with ``A'', ``B'', ``C'', or ``D'' to indicate the correct answer. Your reply should be a single letter and should not contain any additional words.\\
\midrule
\end{longtable}

\paragraph{Consistency evaluation}
The model's responses for the benchmark data are standardised and mapped onto the corresponding class label.
Standardisation involves lower-casing, removing surrounding whitespaces and punctuations.
The model generally conforms to the instruction and responds only with the correct answer.
However, if necessary, additional words are also removed (automatically).
For example, if the model replies ``The answer is `yes'.'' in English and ``Ja'' in German, both responses will be standardised (``yes'', ``ja'') and then mapped onto the corresponding class labels, $l(r)=1$ and $l(r^*)=1$.
Consistency (see \cref{subsec:overall-procedure}, step 4) is then calculated as
\begin{equation*}
    C(R, R^*) = \frac{1}{n} \sum_{i=1}^n \mathds{1}_{l(r_i) = l(r_i^*)}\:,\label{eq:consistency2}
\end{equation*}
where $\mathds{1}$ is again the indicator function.

\subsection{Results}\label{subsec:benchmark_results}

We discuss our results, again starting with accuracy and then continuing with consistency scores.

\begin{figure}[t]
\centering
    \includegraphics[width=0.9\textwidth]{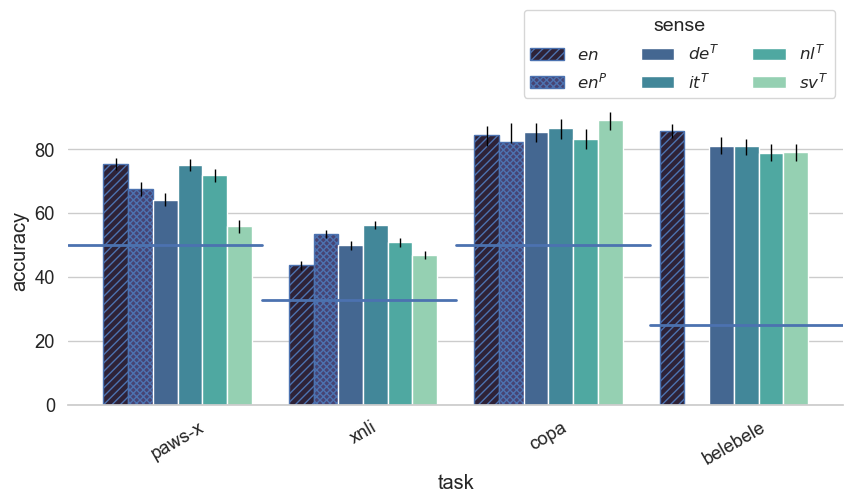}
    \caption{\textbf{Accuracy (\%) for the benchmark datasets, with 95\% confidence intervals.} For Belebele, we have no en$^P$ score, because the model did not provide useable paraphrases. Horizontal lines indicate chance accuracy. Numerical scores can be found in \cref{tab:benchmarks-accuracy}.}\label{fig:benchmarks-accuracy}
\end{figure}

\paragraph{Performance}
We plot the accuracy scores in \cref{fig:benchmarks-accuracy}; horizontal blue lines indicate chance accuracy.
We excluded the results for paraphrases of Belebele, because the model consistently failed to paraphrase this task -- sometimes it ignored the text passage and sometimes it answered the question instead of paraphrasing.
The accuracies for COPA and Belebele are relatively high ($\ge 79$\%) across senses, followed by PAWS and then XNLI.
Performance on Belebele is particularly high, considering there are four answer possibilities, compared to three for XNLI, and two for COPA and PAWS.
Performance on XNLI is particularly low, raising the question of whether this task is perhaps simply not suited for zero-shot evaluation.
Looking into the task in more detail, we suggest that the task may be very prompt-sensitive, with different preferences in different model versions.
For instance, we observed much higher performances with an older gpt-3.5-turbo snapshot as well as GPT-4 on this task.
This may indicate that XNLI is a task that is particularly form-tied, making it an interesting candidate for evaluating multisense consistency.
Overall, we observe that for each task, performance can vary strongly across senses, with up to 19.7\% points on PAWS and up to 12.7\% points on XNLI.

\begin{figure}[ht]
\centering
    \includegraphics[width=0.9\textwidth]{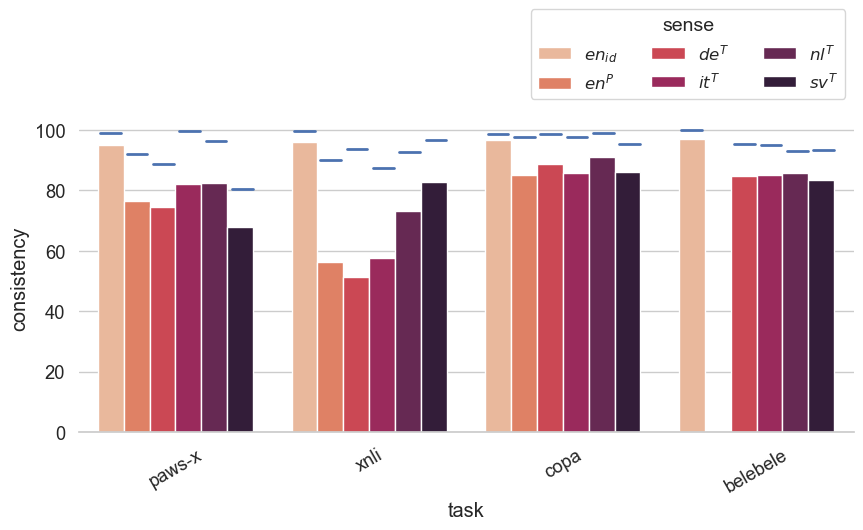}
    \caption{\textbf{Consistency scores (\%) for the benchmark datasets.}
    None of the consistencies between original and alternative sense are close to the maximum possible given the difference in accuracy between the two senses (indicated by the horizontal blue lines), indicating that the models are inconsistent even beyond those differences. Numerical scores can be found in \cref{tab:benchmarks-consistency}.}
    \label{fig:benchmarks-consistency}
\end{figure}

\paragraph{Consistency}
Next, we look at the consistency.
We plot the results in \cref{fig:benchmarks-consistency}, again against the $en$ same-sense baseline ($id$).
Horizontal blue lines indicate the maximal possible consistency when accounting for differences in accuracy.
Overall, model consistency is much lower on some tasks than on others.
With regard to the accuracy scores above, the model tends to be more consistent on tasks it can solve well.
For example, consistency is as low as $51.2$\% on the German translation of XNLI whereas it is above 84\% for all task versions of COPA.
This is not entirely unsurprising because the model can also be consistent when it has a form-dependent task understanding but has learned to generate the correct response for each form (separately).
If the model makes a mistake, however, it is much less likely that it will generate the same mistake in another form, if the generated responses are form-dependent.
The fact that the model overall has a higher consistency on tasks with higher accuracy thus suggests that at least part of its consistency is not due to a form-independent understanding of meaning.
We further investigate this difference in \cref{subsec:conditional_consistency}.
We also see that consistency can vary strongly between senses, ranging from $51.2$\% to $82.8$\% on XNLI, and $67.9$\% to $82.4$\% on PAWS.
A comparison against the baseline scores confirms that inconsistencies go beyond stochasticity inherent to the model.
Considering the results for both \sfd and benchmark data, it seems that accuracy and consistency tend to decrease slightly from higher- to lower-resource languages.
Given that this effect is small, most of the inconsistencies are likely not driven by the choice of senses or the process of generating these senses with the model (see \cref{subsec:sense_quality} for a detailed analysis).
In sum, the systematic benchmark evaluation provides evidence across larger and more diverse datasets than the \sfd evaluation.
The results are in line with our earlier observation that GPT-3.5 is not very self-consistent.


\section{Analysis}\label{sec:analyses}

The results in the previous sections suggest that the meaning representations of the model we investigate are strongly tied to form.
The main evidence for that is the model's inconsistencies across senses.
In this section, we aim to better understand when and why inconsistencies arise.
More specifically:
\begin{enumerate}\setlength\itemsep{0.5em}
    \item We evaluate whether inconsistencies stem from the model's inability to generate meaning-preserving senses, that is, it does not have the ability to adequately paraphrase or translate (\cref{subsec:sense_quality}).
    \item We evaluate whether the model is inconsistent in its task interpretation, in its task execution, or in both (\cref{subsec:interpretation_versus_execution}).
    \item We evaluate consistency conditioned on correctness of the model's responses, because -- as we argue below -- consistently incorrect responses provide stronger evidence for a form-independent task understanding than consistently correct ones (\cref{subsec:conditional_consistency}).
    \item We study if there is a connection between requested information and prompt language that could provide direct evidence for form dependency as a source of inconsistency (\cref{subsec:direct-evidence}).
\end{enumerate}
We present these analyses for the simple facts, the benchmarks, or both, as appropriate.


\subsection{Quality of alternative senses}\label{subsec:sense_quality}

The metric we propose conflates task understanding of the ``primary'' sense and ability to generate different senses: if a model is not able to generate adequate translations or paraphrases, this may give rise to inconsistencies even if it has a form-independent understanding of meaning.
While both are important qualities, and the metric favours models that do well across the board, it makes sense to consider the two parts separately as well.
Differences in task understanding for high-quality senses point to a form-dependent task understanding whereas, as pointed out earlier, a failure to translate or paraphrase may not.
For example, while a poor task understanding can lead to a bad translation, a poor translation might also arise from a poor command of the target or source language, or an inability to translate.
To examine if inconsistencies are due to one of the latter causes, we investigate the quality of the paraphrases and translations.

\paragraph{Translation and paraphrase quality}
First, we check the quality of the translations and paraphrases for both \textsc{simple facts} and benchmark data.
To evaluate the instruction data, we ask native speakers of each language, who are also fluent in English, to verify whether the paraphrases and translations are correct and meaning-preserving.
For the \sfd data, we consider the templates; for the benchmark data, we consider the task instructions (see Appendix~\ref{app:C_instructions} for a full list of these).
For both types of data, the instructions were largely judged to be grammatically correct and meaning-preserving, although they tend to stay relatively close to the English original, such that a native speaker might prefer a slightly different wording.

Next, we automatically evaluate whether the numbers for the \textsc{arithmetics} task are translated correctly.
Each datapoint consists of a pair of numbers (see \cref{tab:templates}) and the translation counts as correct if both numbers are translated correctly.
We find that the translations are highly accurate for German ($99.6$\%) and Dutch ($99.4$\%), but less so for Italian ($89.2$\%) and Swedish ($81.0$\%).
Still, the proportion of wrong translations is significantly smaller than the proportion of inconsistencies across all languages, and can thus explain only a small part of the inconsistencies for that task.

For the benchmark data, we further evaluate the quality of the translations of the \emph{task input data}, by comparing them to reference data, available either in the benchmark itself (in case of Belebele) or in the multilingual benchmark versions we use.
We report BLEU~\citep{papineni-etal-2002-bleu}, ROUGE~\citep{lin-2004-rouge} and COMET-22~\citep{rei-etal-2022-comet} scores, all commonly adopted measures of translation quality, in \cref{tab:translation_quality}.
All metrics indicate that the model's translations are of high quality across tasks and languages.
The high scores suggest that, for most of the considered source-target language combinations, inconsistencies can largely not be ascribed to changes in meaning induced by the translation.

\begin{table}[]
\caption{\textbf{Translation quality.} We consider the quality of the translations of the input data to different senses, according to different commonly used metrics. All scores are comparatively high, suggesting that the model's inconsistencies are not driven by an inability to translate.}\label{tab:translation_quality}
\centering
\begin{tabular}{l|l||c|c|c|c|c}
\toprule
{}         & {} & \textbf{bleu} & \textbf{rouge1} & \textbf{rouge2} & \textbf{rouge-l} & \textbf{comet-22} \\
\midrule
\midrule
paws                        & de$^T$      &   57.5    &   0.81    &   0.65   &   0.77   &  0.85 \\
\midrule
xnli                         & de$^T$     &   41.9    &   0.69    &   0.49   &   0.66   &  0.84 \\
\midrule
copa                        & it$^T$      &   40.9    &   0.66    &   0.45   &   0.64   &  0.86 \\
\midrule
\multirow{4}{*}{belebele}   & de$^T$      &   41.1    &   0.69    &   0.46   &   0.63   &  0.84 \\
                            & it$^T$      &   38.1    &   0.69    &   0.44   &   0.61   &  0.85 \\
                            & nl$^T$      &   34.3    &   0.68    &   0.40   &   0.57   &  0.85 \\
                            & sv$^T$      &   44.0    &   0.73    &   0.53   &   0.68   &  0.86 \\
\bottomrule
\end{tabular}
\end{table}

\paragraph{Translation quality vs consistency}
To investigate the relationship between translation quality and consistency in more detail, we run several follow-up analyses.
First, we calculate the Pearson correlation between consistency and COMET scores.
The correlation for XNLI is negative ($\rho = -0.06$), and for COPA ($\rho = 0.07$) and PAWS ($\rho=0.11$) it is relatively small.
For Belebele the correlations are also rather small ($\rho$ between $0.08$--$0.13$) with a somewhat higher value for Swedish ($\rho=0.21$).
Second, we evaluate the consistencies for a subset of the best translations, considering only datapoints with COMET scores greater than $0.80$.
Relative to the original scores across all datapoints, consistency scores change between $-2.7$ and $2.0$ percent points across datasets and languages; based on a two-sided t-test this difference is not significant ($p>0.9$).
Finally, we evaluate the model's consistency when replacing the self-translated input data with the ground truth references for each language.
When reference data is available, we pair the model's translation of the instruction with the benchmark data for the corresponding target language (e.g. $de^T$ instruction and $de$ input data).
It turns out that the model's consistency \emph{decreases} in six out of seven cases (by up to $-5.2$\%) and increases in one case (by $0.7$\%).
In other words, the model tends to be more consistent when the alternative sense is self-generated.
This result also highlights the importance of using the model's own translations and paraphrases: despite imperfect translations and paraphrases, the model treats self-generated senses as slightly more meaning-equivalent than externally generated ones.
These additional analyses show that translation quality can affect consistency but is not a major driver of the inconsistencies observed in our experiments.

\subsection{Interpretation versus execution}\label{subsec:interpretation_versus_execution}

Next, we investigate if, when a model is inconsistent across senses, this inconsistency stems from an inadequate understanding of what the task is or from an inadequate execution of that task in that specific language.\footnote{
This distinction is related to the fact that we evaluate the model's understanding with different \emph{tasks}.
Based on Frege's observation that different senses can have the same meaning, we need to create an interface that allows us to test whether LLMs actually assign the same meaning to different senses.
In our case, this interface consists of the task that the model is supposed to carry out on a given input.
Thus, the analysis can also be considered a way to disentangle the model's meaning understanding of the input sentences from its meaning understanding of the instructions.
}
To exemplify this, compare the scenario in which you are asked to judge whether one English sentence implies the other, but the request is made in a language that you do not have a great command of with the scenario in which the question is asked in English, but the sentences to be judged are in a language you do not understand well.
Because the \sfd does not have separate instruction and task data, we analyse this only for the benchmark data.

To disentangle the impact of changing the sense of the task instruction and the task input data, we run an ablation experiment.
Specifically, we assess the model's consistency when paraphrasing/translating only the instruction while keeping the original input data (condition $I$), as well as its consistency when paraphrasing/translating only the input data while keeping the original instruction (condition $X$).
The resulting consistency scores are displayed in \cref{tab:consistency-ablation} and the corresponding accuracies in Appendix~\ref{app:F_accuracy_ablation}.
Neither consistencies for translating only the instructions nor those for translating only the input data are at their maximum, indicating that the model is inconsistent in both interpretation and execution.
Whether inconsistencies in execution or interpretation are more pronounced depends largely on the task.
In particular for XNLI, where the instruction is very complex, consistencies are higher when using the same instruction compared to using the same input data.
For tasks with comparatively simple instructions, the pattern is at least partially reversed.
Consistency is always lower when using the same instruction but different input data for Belebele and COPA, and in some cases also for PAWS.
When paraphrasing/translating both instructions and input data (cf. \cref{fig:benchmarks-consistency} / \cref{tab:benchmarks-consistency}) consistencies are mostly lower than for either ablation.
Thus, inconsistencies seem to be driven by differences in both task interpretation and execution.
Differences in execution are more pronounced unless the task is difficult to interpret.

\begin{table}[]
\caption{\textbf{Consistency scores (\%) for the ablation experiments.}
We analyse whether consistencies mainly arise from differences in task interpretation or execution, by considering ablations in which we translate/paraphrase only the instruction (columns $I$) or only the input data (columns $X$).
Where inconsistencies are more pronounced depends largely on the task.
Mostly for XNLI, interpreting the (comparatively) complex instruction appears to be more challenging than understanding the sentence.
    }\label{tab:consistency-ablation}
\centering
\begin{tabular}{l|c|c|c|c|c|c|c|c}
\toprule
{}          &   \multicolumn{2}{c|}{paws} & \multicolumn{2}{c|}{xnli} & \multicolumn{2}{c|}{copa} & \multicolumn{2}{c}{belebele} \\
\midrule
{}          & $I$ & $X$ & $I$ & $X$ & $I$ & $X$ & $I$ & $X$ \\
\midrule
en$^P$      &   89.5  & 78.4      & 64.0  & 86.7      & 90.2 & 87.0       &   -    & 94.4  \\
de$^T$      &   77.8  & 81.1      & 57.9  & 88.5      & 94.0 & 88.6       &   94.1 & 84.7  \\
it$^T$      &   91.2  & 82.0      & 60.9  & 88.9      & 91.8 & 86.2       &   94.4 & 83.3  \\
nl$^T$      &   86.4  & 83.3      & 77.9  & 88.6      & 93.2 & 90.0       &   94.1 & 86.7  \\
sv$^T$      &   72.7  & 80.3      & 82.4  & 88.6      & 91.0 & 87.4       &   94.2 & 84.9  \\
\bottomrule
\end{tabular}
\end{table}


\subsection{Consistency vs correctness}\label{subsec:conditional_consistency}

We further investigate if there is a difference in consistency between examples for which the model provides a correct answer and those for which it provides an incorrect answer.
This comparison is interesting because correct and incorrect consistent examples provide different levels of evidence for consistency of meanings beyond form.
If a model gives consistently correct answers for an example, it is possible that it has inferred those correct answers independently from the data for the respectively languages. 
In that case, consistency does thus not necessarily point to a form-independent understanding of the particular question.
This is much less likely the case for \emph{incorrectly consistent} examples, as it would require that the data the model was trained on contained the same error for both languages.
Being consistently incorrect across two examples thus points to an error in the model's understanding but provides stronger evidence for the consistency of its underlying representations than examples that are consistently correct.

\begin{figure}[]
\centering
    \begin{subfigure}[b]{0.49\textwidth}
    \includegraphics[width=\textwidth]{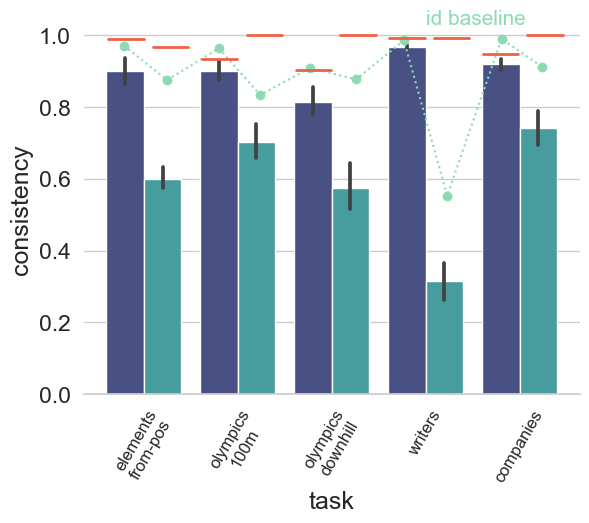}
    \caption{simple facts}\label{fig:conditional_consistency_facts}\hfill
    \end{subfigure}
    \begin{subfigure}[b]{0.49\textwidth}
    \includegraphics[width=\textwidth]{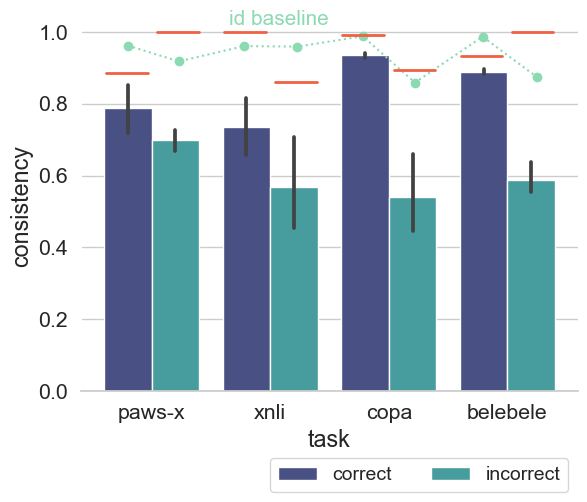}
    \vspace{-0.05cm}
    \caption{benchmarks}\label{fig:conditional_consistency_benchmarks}\hfill
    \end{subfigure}
    \caption{\textbf{Consistency scores conditioned on correctness.} Error bars indicate 95\% confidence intervals. Examples that are consistent and \emph{incorrect} provide stronger evidence for a form-independent meaning understanding than \emph{consistent correct} examples, because it is less likely that incorrect information was inferred independently. The large difference between consistent correct and consistent incorrect in this plot thus indicates that -- likely -- some of the consistent correct examples were correct independently. As in the previous plots, upperbound consistency given the individual sense accuracies is given by horizontal lines. The dotted line indicates the $id$ baseline (two runs in English).
    }\label{fig:conditional_consistency}
\end{figure}

\cref{fig:conditional_consistency} shows the consistency scores conditioned on whether the model was correct on the source task ($en$), for both the \textsc{simple facts} (left) and the benchmark data (right).
The scores are averaged across senses and the $id$ baseline is given by the dotted lines.
We can see that the model is always more consistent on correct responses than on incorrect responses, suggesting that the its responses are -- at least in some cases -- consistent simply because they are independently correct in both languages.
Given that the difference between the two conditions (correct vs incorrect) is more pronounced for different senses than the same-sense baseline, it cannot solely be attributed to stochasticity for cases where the model's distribution is relatively flat among the highest scoring answers but it can nevertheless not answer ``I don't know''.
In conclusion, not only when answering simple factual questions, but also across a range of NLU tasks the model seems to infer a significant amount of its responses separately for each sense.

\subsection{Direct evidence for form-dependency}\label{subsec:direct-evidence}

The analyses above all provide converging but indirect evidence for form dependencies in the model's understanding.
In this final analysis, we aim to establish a direct connection between the type of information the model is asked about and the form of the question.
It is plausible that certain information is more often presented to the model in a certain form during training.
For instance, information about Italian companies likely occurs more often in Italian text than in Swedish text and vice versa.
If acquired meanings transcended the form they were acquired in, this should not matter: once acquired, a fact should be accessible in whatever languages mastered by the model.
Thus, if a model scores comparatively better in the language that is related to the information requested, this points to a form-dependent question understanding.
To test this hypothesis we exploit the controlled structure of the \textsc{writers} and \textsc{companies} datasets.
Both datasets comprise five subsets of equal size (see \cref{tab:templates}).
Each subset contains facts that can be considered somewhat specific to one of our test languages, establishing two conditions of matching or mismatching prompt language and target information.
Accordingly, we investigate whether prompting the model in the information-specific language yields higher accuracy compared to prompting it in another language.

\begin{figure}[]
\centering
    \begin{subfigure}[b]{0.49\textwidth}
    \includegraphics[width=\textwidth]{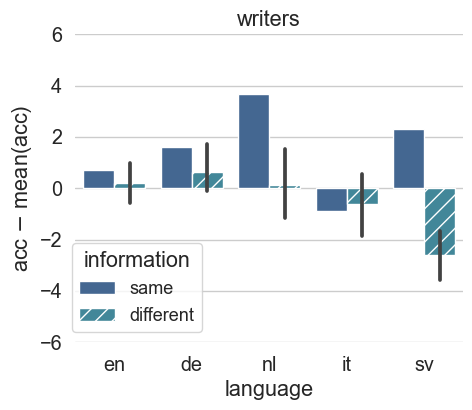}
    \end{subfigure}
\hfill
    \begin{subfigure}[b]{0.49\textwidth}
    \includegraphics[width=\textwidth]{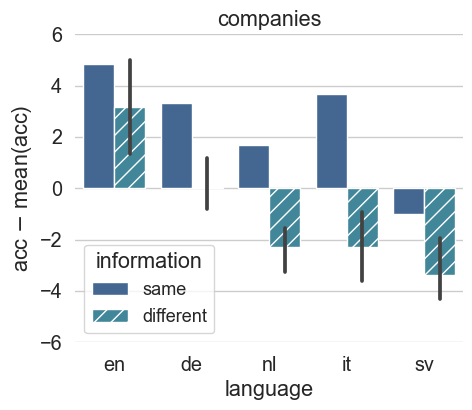}
    \end{subfigure}
\caption{\textbf{Language-dependent knowledge for the \sfd dataset.} Error bars indicate 95\% confidence intervals.
    For each language, we compute how its accuracy when asked about information matching that language compares to its accuracy when asked about information not matching that language (e.g.\ asking about Dutch writers in Dutch vs in Swedish), compared to the overall averages for those groups.
Generally, the model has higher accuracy when the prompt language and requested information pertain to the same country (plain bars) than when it is asked about non-matching information (hatched bars).
}\label{fig:language-specific-knowledge}
\end{figure}

In Figure~\ref{fig:language-specific-knowledge}, we plot i) the absolute difference between the accuracy of the model when prompted in the language \emph{matching} the data subset (e.g.\ asked about Dutch writers in Dutch) and the overall average accuracy for all languages on that subset, and ii) the absolute difference between the accuracy of the model when prompted on \emph{mismatched} subsets (e.g.\ asked about non-Dutch writers in Dutch) and the overall average for all languages for that same group of subsets.
With the exception of Italian on the \textsc{writers} task, the model is always comparatively (and sometimes absolutely) better on the language-matched subsets (plain blue bars) than on the mismatched subsets (hatched turquoise bars).
For example, when prompting the model in Dutch on the Dutch \textsc{writers} subset, accuracy is almost 4\% higher compared to the average accuracy for this subset across prompts (including $nl$).
A two-sided t-test between the deviations from the mean for cases with matching versus mismatching information and prompt languages is highly significant ($p=0.001$).
While this analysis covers only two datasets, the results provide direct, positive evidence for a form-dependent task understanding.

\section{Related work}\label{sec:related_work}

In this work, we considered LLMs as explanatory models of meaning.
Here, we discuss work related to the various aspects of our study.
In particular, we discuss studies that have used LMs as explanatory models of language or language processing (\cref{subsec:llms-as-explanatory-models}); work that explicitly discusses form and meaning in LLMs (\cref{subsec:form-meaning-LLMs}); and studies that have involved (multilingual) consistency in LLM evaluation protocols (\cref{subsec:rw_consistency}).

\subsection{LLMs as explanatory models}\label{subsec:llms-as-explanatory-models}

Despite the many differences between biological and artificial neural networks, the latter have been extensively investigated as explanatory models to further our understanding about human cognition, primarily in the domains of vision and natural language.
In the field of natural language processing, these endeavours have spanned a large range of phenomena and questions.
As some understanding of how neural networks behave or what they represent is a prerequisite for using them as explanatory models, such studies often interweave various interpretability methods with (psycho)linguistic theory.
Here, we focus specifically on studies that use (modern) LMs and make an explicit attempt to reconnect their findings with human processing, linguistics or cognition.\footnote{There also exists quite some literature that aims to directly draw connections between the representations in neural networks and in the human brain. We consider that beyond the scope of this paper, and will not further discuss it.}

\paragraph{Nested hierarchical processing} One subject elaborately explored in linguistically inspired studies of LMs is their ability to process hierarchical structure in language.
Starting from the work of \citeauthor{linzen-etal-2016-assessing} in \citeyear{linzen-etal-2016-assessing}, a wave of studies have considered long-distance subject-verb agreement as a proxy for this ability \citep[e.g.][]{gulordava-etal-2018-colorless,giulianelli-etal-2018-hood}.
The most clear-cut example of using subject-verb agreement in LLMs in an explanatory fashion is the series presented by \citet{lakretz-etal-2019-emergence,lakretz2021cognition} and \citet{baroni2023role}, who used a psycholinguistic experiment to assess whether a mechanism for processing nested dependencies they found in LMs may be deployed by humans as well.

\paragraph{Inflectional morphology} Another topic that has since long been used as a testing ground for answering questions about linguistic generalisation in humans and the viability of neural networks as models of cognition is inflectional morphology.
The amount of literature on this topic is too vast to discuss in detail in this work, for a concise summary, we refer to the related work section of \citet{dankers-etal-2021-generalising}.

\paragraph{Processing difficulty} Lastly, starting from \citet{elman1990finding}, there is a long tradition of trying to link performance of mostly recurrent neural networks to human processing difficulty \citep[i.a.]{christiansen1999toward,frank2011insensitivity,futrell-levy-2017-noisy}.
Several of such studies have considered surprisal (i.e.\ predictive difficulty) to study hypotheses regarding the role of retrieval and prediction in defining human processing difficulty.
Among others, \citet{wilcox2020predictive}, \citet{van2021single} and \citet{huang2023surprisal} show that surprisal in neural networks often differs strongly from human reading-time data, and that predictive difficulty is thus likely insufficient to explain processing difficulty.
In a similar vein, several others have considered how LMs process garden path sentences \citep[e.g.][]{ulmer-etal-2019-assessing,van2018modeling,van2021single,arehalli-etal-2022-syntactic} -- in psycholinguistics often studied to investigate if humans maintain multiple parses at once.
\citet{ryu-lewis-2021-accounting} and recently \citet{timkey2023language} focus more on the retrieval side, and show positive results concerning the similarity of attention head behaviour with effects observed in human experiments.

\subsection{Form and meaning in LLMs}\label{subsec:form-meaning-LLMs}

Currently, the degree to which LLMs can and do have meaning-based, rather than mere form-based knowledge and understanding is widely debated~\citep[e.g.][]{mitchell_krakauer_2023,raji2021everything}.
To begin with, there is no agreement in the community on whether LLMs can \textit{in principle} learn meaning from text.
While some argue that meaning cannot be learned from form alone~\citep[e.g.][]{bender-koller-2020-climbing} others disagree or argue that the training signal for some LLMs goes beyond form~\citep[e.g.][]{piantadosi2022meaning,mollo2023vector,pavlick_2023,mandelkern2023language}.
Importantly, current NLU benchmarks do not provide the means to disentangle the roles of form and meaning~\citep[e.g.][]{heinemanrethinking}.
If a model achieves a high score on a benchmark, it is not clear whether the model relies on specific lexical patterns or general principles when performing the task~\citep[e.g.][]{ray-choudhury-etal-2022-machine}.
In some cases, LLMs have been found to exploit spurious statistical patterns or rely on information memorised from the training, rather than a flexible and generalisable task understanding~\citep[e.g.,][]{geva-etal-2019-modeling, mccoy-etal-2019-right, mckenna2023sources}.
Adversarial datasets~\citep[e.g.][]{nie-etal-2020-adversarial} are designed precisely to expose such shortcut learning behaviours ~\citep[for an overview of shortcut learning, see][]{du2023shortcut}.
Despite this uncertainty, it is common to construct ``understanding'' benchmarks without considering this question.
Instead, ``understanding'' is typically reduced to generalisation across many different tasks~\citep[e.g.][]{wang-etal-2018-glue,wang-2019-etal-superglue,hendrycks2021measuring}.
An evaluation of consistency can also be considered a generalisation evaluation.\footnote{See Appendix \ref{app:A_genbench} for a GenBench eval card~\citep{hupkes2023stateoftheart} that classifies our work in the context of generalisation research.}
However, by evaluating a model across different senses with the same meaning (i.e.\ different versions of the same task) rather than different meanings (i.e.\ different tasks), it is possible to uncover form dependencies that stand in contrast to a human-like task understanding.

\subsection{Consistency}\label{subsec:rw_consistency}

Various studies have shown that inconsistencies are common in LLMs (and have suggested methods for improving consistency, which is not our focus).
To begin with, investigations of model robustness have revealed that even minor (meaning-preserving) perturbations of the model input can strongly affect the generated output~\citep[e.g.][]{chakraborty-etal-2023-zero,weber-etal-2023-mind,wang2023large,mizrahi2023state,Podkorytov_2021}.
Other than that, studies are mostly concerned with self-consistency in natural language inference (NLI)~\citep[e.g.][]{minervini-riedel-2018-adversarially, wang2018simply, li-etal-2019-logic, hosseini-etal-2021-understanding} and question answering~\citep[e.g.][]{kassner-schutze-2020-negated, alberti-etal-2019-synthetic, mitchell-etal-2022-enhancing, chen-etal-2021-nli-models, elazar-2021, kassner-etal-2021-beliefbank, asai-hajishirzi-2020-logic, hosseini-etal-2021-understanding}.
For example, \citet{kassner-etal-2021-beliefbank} created a dataset of sentence pairs that are subject to certain constraints (e.g.\ if \textit{X is a dog} is true, \textit{X has a tail} must also be true).
Their evaluation of \textit{Macaw}~\citep{macaw_model}, a fine-tuned T5 model, revealed significant inconsistencies in the model's beliefs.
In the same vein, various GPT models fail to generalise from statements of the form ``A is B'' to ``B is A''~\citep{berglund2023reversal}.
More similar to our work, \citet{elazar-2021} studied whether factual knowledge in masked LMs is invariant to paraphrasing.
To this end, they created \textsc{ParaRel}, a dataset containing cloze-style English paraphrases (e.g. \textit{Homeland originally aired on [MASK]}, \textit{Homeland premiered on [MASK]}), which was, for example, recently used to reveal inconsistencies across various LLaMA~\citep{touvron2023llama} and Atlas~\citep{atlas_JMLR:v24:23-0037} models~\citep{hagstrom-etal-2023-effect}.
In the studies mentioned here, consistency is either evaluated against a network of logical relationships between beliefs or by generating different forms of the same meaning through paraphrasing.
BECEL~\citep{jang-etal-2022-becel} is a benchmark for evaluating these two types of consistency (logical and semantic) across various tasks.
For each task, the benchmark provides an alternative version (e.g.\ for semantic consistency the inputs are paraphrased) to compare the model's answers across task instances.
This benchmark has recently been used to evaluate ChatGPT, showing that it is more consistent for negations than other LLMs, but still likely to generate different responses to paraphrases of the same meaning~\citep{jang2023consistency}.
Except for \citet{jang2023consistency} and our own preliminary work~\citep{ohmer_2023} consistency usually relies on different forms of the same meaning that are generated externally from the model.
We focus on true \emph{self}-consistency, where alternative senses are generated by the model under investigation, to ensure that the model -- if it can assign meaning -- should assign the same meaning to the original and the derived sense.

\paragraph{Multilingual consistency}
Given that we generate different forms through translation our approach is related to multilingual model evaluation.
Multilingual benchmarks are usually generated from existing benchmarks through expert translations~\citep[for a more expansive overview, we refer to][Appendix D]{hupkes2023stateoftheart}.
Prominent examples include PAWS-X~\citep{yang-etal-2019-paws}, XCOPA~\citep{ponti-etal-2020-xcopa}, and XNLI~\citep{conneau-etal-2018-xnli}.
Furthermore, multilingual tasks have been combined to form multilingual multitask benchmarks~\citep[e.g.][]{hu-etal-2020-xtreme,ruder-etal-2021-xtreme,liang-etal-2020-xglue}. 
All of these benchmarks reveal language-dependent differences in performance for current multilingual LLMs, which indicates that the models' responses to the original and the translated task versions are not perfectly consistent.
Recently, \citet{qi2023crosslingual} combined consistency and multilingual evaluation by introducing a ranking-based consistency metric for evaluating knowledge consistency across languages independently from accuracy.
They found that consistency correlates strongly with the sub-word vocabulary overlap between two languages, suggesting that knowledge transfer between languages relies on shallow features rather than a true understanding.
In contrast to existing multilingual evaluation approaches, we aim to evaluate self-consistency by detecting language-dependent changes in model responses, relying on the model's own translations.


\section{General discussion}\label{sec:discussion}

In this paper, we proposed a paradigm to investigate whether LLMs acquire form-independent notions of meaning, with the larger aim of assessing the viability of using them as explanatory models to better understand the concept of meaning.
In this last section, we summarise the key aspects of our approach and the main findings from our experiments (\cref{subsec:summary}), discuss the separation of form and meaning in humans versus LLMs in light of our findings (\cref{subsec:form-and-meaning-humans-vs-llms}), and revisit the discussion on using LLMs as explanatory models of meaning, specifically considering the role of multisense consistency therein (\cref{subsec:llms_as_explanatory_models_of_semantic_understanding}).

\subsection{Summary}\label{subsec:summary}

Motivated by the successes of LLMs as explanatory models of \emph{form}, we are interested in their potential as explanatory models of \emph{meaning}.
Our analysis takes inspiration from philosophy of language.
Based on Frege's distinction between sense and reference, we propose a paradigm to study if LLMs, trained on only forms, possess form-independent notions of meaning.
Specifically, we evaluate the self-consistency of a model across different meaning-preserving forms (senses), generated by the model itself.
The main idea underpinning this paradigm is that if a model's understanding extends beyond form, it should produce consistent responses to different senses that express the same meaning -- provided it understands the equivalency between these different senses.

Using this paradigm, we investigated the form-dependency of natural language understanding in GPT-3.5, a state-of-the-art language model.
We conducted experiments with a novel benchmark with simple factual questions and different NLU benchmarks.
The former provides unambiguous evidence of form dependency, while the latter speak to the extent of this form dependency across various NLU tasks.
We detected inconsistencies for all tasks, across all generated senses, both in paraphrases and translations.
Our analyses control for explanations other than a form-dependent understanding: inconsistencies are neither due to inherent stochasticity, 
nor due to changes in meaning in the sense-generation process.
They also help us better understand the nature of the model's inconsistencies, by showing that the model is inconsistent in task interpretation and execution and that the inconsistencies are more pronounced in incorrect examples than in correct examples.
These findings indicate that the model infers its responses separately for each sense and highlight the limitations of current LMs in capturing the true nature of meaning.

\subsection{Form and meaning in humans versus LLMs}\label{subsec:form-and-meaning-humans-vs-llms}

Form-independent meaning is critical to human understanding.
Many tasks that we encounter share a common abstract structure.
In solving familiar and novel tasks we can exploit this structure by accessing the same knowledge, reasoning process, or skill~\citep[e.g.][]{tenenbaum_2011, barsalou_abstraction_2005, gentner_analogical_inference}.
Furthermore, neurological evidence supports that the brain maintains abstract task representations which are used in generalisation~\citep[e.g.][]{Liu2019-gh, McKenzie2014-xx, Badre2018-gp, vaidya_10.7554/eLife.63226}.
In our implementation, different forms of the same task correspond to different languages or paraphrases.
Also for this specific instance, there is evidence for a form-independent understanding in humans.
Studies with bilinguals and second-language learners collectively support the view that lexical-level representations (form) are independent whereas semantic-level representations (meaning) are shared~\citep{kroll2014lexical, Hernandez2005-qx, francis_2005}.
The multilingual inconsistencies observed in our experiments with ChatGPT suggest that the model does not possess such form-independent semantic-level representations.
Further evidence for a form-dependent task understanding in LLMs comes from multilingual consistency evaluations with model-external translations.
While these experiments do not guarantee that the different translations are meaning-equivalent according to the model, they still indicate that LLM responses seem to be largely driven by the lexical form of the input~\citep{qi2023crosslingual}.

To different degrees, both translations and paraphrases preserve the meaning of the original expression.
In our work, we tested both translating and paraphrasing as sense-generation methods.
However, translation equivalents and synonyms are treated differently in human cognition.
For example, monolingual and bilingual children accept two names for the same object -- violating the mutual exclusivity assumption -- if the two names come from distinct languages but not if they come from the same language~\citep{au_glusman_1990}.
In particular, it seems that translation-equivalents have a closer cognitive status than within-language synonyms~\citep{francis_2005}.
The model's consistency for translations versus paraphrases stands in contrast to the empirical evidence that changes in language have a more similar cognitive role than changes in wording.
If anything, consistency tends to be higher for English paraphrases than translations (see for example \cref{tab:simple-facts-consistency}).
In conclusion, LLMs do not seem to separate between form and meaning in the way humans do.

It is important to keep in mind that looking up a fact with an LLM is not as straightforward as looking up a fact in an encyclopaedia.
Our experiments show that LLM responses to factual questions may vary between different representational forms of the same input, even if the model judges these forms to be meaning-equivalent.
LLMs might (at least partially) lack an anchor for the linguistic forms they encounter, which humans naturally find in the physical world and social interactions~\citep{bisk-etal-2020-experience}.
Their responses, especially to factual questions, should thus be considered with caution and users should be aware that other knowledge sources are more reliable.
\citet{chang2023language} suggest that many weaknesses of LLMs, including form-dependencies, can be framed as under- and over-generalisation errors.
When a model is sensitive to small, meaning-preserving changes to the input, when recalling facts, this can be considered an under-generalisation of the underlying factual knowledge.
The model may compensate for this failure by over-generalising other patterns, thus falling back on certain heuristics to generate an answer.
In general, it is important to keep in mind that LLMs and humans are shaped by different pressures when making a comparison.
For example, while LLM accuracy is strongly influenced by the probability of the task to be performed, the probability of the target output, and the probability of the provided input, humans are likely better at generalising their task understanding across such variations~\citep{mccoy2023embers}.

\subsection{LLMs as explanatory models of meaning: The role of multisense consistency}\label{subsec:llms_as_explanatory_models_of_semantic_understanding}

What are the consequences of our findings for the role of LLMs as explanatory models of semantic understanding in humans?
Up until now, the discussion has largely revolved around their capacity to represent symbolic structure and to capture the nature of language use, including communicative intent and grounding in the world.
While there are a priori arguments that LLMs fail at both these fronts, let us consider some arguments in favour of such capacities.
Concerning symbolic structure, arguments come, for example, from interpretability studies that identified dedicated neurons for encoding specific knowledge~\citep{dai-etal-2022-knowledge}, concepts~\citep{geva-etal-2021-transformer}, or skills~\citep{wang-etal-2022-finding-skill} in transformer-based LMs.
Concerning perceptual grounding, it has been argued that important aspects of meaning are captured by the role a certain concept plays, that is, how it relates to other concepts within a representational framework, rather than being defined by an external referent~\citep{piantadosi2022meaning}.
When studying the internal representations of LLMs, the organisation of concepts -- measured through similarity relationships -- indeed seems to match the ground-truth organisation of perceptual concepts such as colours~\citep{abdou-etal-2021-language} or spatial relations~\citep{patel2022mapping}.
The lack of self-consistency revealed by our findings opens up a new dimension to be considered when making such arguments.
For example, it is not only relevant whether LLMs can encode symbolic structure and whether they encode concepts in line with a human-like conceptual structure, but also whether these encodings are consistent across senses.
In other words, to establish a strong correspondence between LLM and human concept encodings, these encodings should bear resemblance across different senses.

With that, we believe that measuring multisense consistency could be a useful addition to the toolkit used to evaluate the extent to which models can understand natural language.
The method can be used to assess generalisation ability beyond specific forms.
It offers affordability and applicability to different evaluation tasks, while also mitigating the risk of evaluating on data that the model has already encountered during training.
As such, multisense evaluation could serve as a complement to performance-based model evaluation.
Reporting consistency next to standard evaluation metrics like accuracy, BLEU, or F1-scores will make model evaluation more meaningful in providing an estimate of how well the model understands a given task beyond its specific form.
Our paradigm can be cheaply and easily expanded to include more languages, tasks, models, and notions of ``sense''.
Our choice to generate senses through translation is well-suited for evaluating current and future models, given the growing trend towards multilingual models with increasingly proficient translation abilities.
Nevertheless, numerous other multisense evaluations are conceivable.
For instance, senses could be generated through various word- and sentence-level perturbations~\citep[e.g.][]{wang2021adversarial}, across accents or dialects, or across different modalities.
Last but not least, calculating consistency for various tasks may help disentangle ``unfounded'' language-specific differences (forming the focus of our analysis) from differences related to cultural bias.
Therefore, we encourage other researchers to treat multisense consistency as an integral part of benchmarking.

The consistency evaluation is only interesting if the model does not master the task on each sense, in which case its responses are trivially consistent.
Although it is usually impressive when a model achieves high scores on a benchmark that was challenging for the previous model generation, the community rarely concludes that this model has mastered the skill this benchmark is supposedly testing.
As a result, benchmarks are usually replaced by more challenging successors when this happens.
Thus, we think it is likely that challenging benchmarks, which can be used for a non-trivial consistency evaluation will continue to be available.
Still, it is important to mention that consistency should be evaluated in experiments where the main source of potential inconsistencies is form-dependency.
Model mistakes and inconsistencies should not be enforced on purpose, for example through ambiguous instructions.
Further analyses, such as controlling the quality of the generated senses or calculating the proportions of consistent correct versus incorrect responses (see \cref{sec:analyses}), can help to rule out alternative explanations.

Crucially, multisense consistency experiments can primarily provide \emph{negative evidence}.
After all, even if an LLM is perfectly self-consistent, it could be mastering each form independently without relying on a shared meaning.
With that, our method can be grouped together with other methods probing for human-level understanding that, when successfully passed, provoke thought about what ``human-level understanding'' means, rather than providing a proof for it~\citep[e.g.][]{biever_2023,replace_turing_test_2023}.


\section*{Acknowledgments}
We would like to thank the anonymous reviewers and Mortimer von Chappuis for their detailed and helpful feedback on our first submission.
We would also like to thank Marco Baroni and Ryan Nefdt for their valuable feedback on this project at an earlier stage.
Finally, we would like to thank Henrik Löfberg for his help with the evaluation of the Swedish translations.


\bibliography{custom,anthology}


\newpage
\begin{appendices}

\section{Simple facts datasets}\label{app:simple_facts}

We use five different datasets to test for factual knowledge.
To facilitate the dataset curation, we focused on facts that are usually presented in a table format and can be queried with the same template question regardless of the exact datapoint.
At the same time, we tried to cover different domains of factual knowledge, including arithmetics, science, sports, economy, and literature.
Note that these datasets are not intended to serve as fully fledged benchmarks of factual knowledge but rather as a proof-of-principle.
In the following, these datasets are described in detail.
We describe only the base data.
The corresponding instructions are given in \cref{tab:templates} in the main text.
The csv files for each dataset can be found in our repository.

\subsection*{Arithmetics}

The \textsc{arithmetics} dataset tests for the sum of two numbers.
The two numbers are sampled randomly between $1$--$1000$ and to make the questions more different between languages, we chose to spell out the numbers in words.
We wrote functions to map numerals to spelled-out numbers in all the languages we consider (see our repository).
The function for English was used to generate the original dataset once the integers were sampled.
The functions for the other languages were used to evaluate whether the model correctly translated the English (spelled-out) numbers when generating other senses.
The model, in turn, is asked to reply in numerical form, such that the answers can easily be validated.
For instance one datapoint could be $d = ($ \textit{five hundred seventy-three}, \textit{twenty-seven} $)$ and the corresponding set of correct answers would be $A_d=\lbrace 600\rbrace$.
We sample 500 pairs of numbers, giving us a total of 500 datapoints.

\subsection*{Elements}
The \textsc{elements} dataset tests for the atomic number of chemical elements.
Each datapoint consists of a chemical element (denoted by its element symbol), as well as its position on the periodic table (given by period and group).
For example, Helium, which is in period 1 and group 18, is given by $d=(\text{\textit{He}}, 1, 18)$.
The dataset is used for two different tasks.
In the \textsc{from-element} subtask, the atomic number of an element has to be determined from its chemical symbol.
In the \textsc{from-position} subtask, the atomic number of an element has to be determined from its position in the periodic table.
Hence, in both cases, the set of correct answers for the above datapoint is $A_d=\lbrace 2\rbrace$.
The model is instructed to reply with the correct number allowing for easy evaluation against the ground truth.
We ignore the f-block of the periodic table, resulting in a total of 90 datapoints (per subtask).

\subsection*{Olympics}
The \textsc{olympics} dataset tests for the names of Olympic medallists.
It is used for two subtasks.
The \textsc{100m} subtask asks for the medallists in the $100$m competition (Summer Olympics).
The \textsc{downhill} subtask asks for the medallists in the downhill competition (Winter Olympics).
Information on the medallists for these competitions can be found on various sites on the internet, e.g. \url{https://olympics.com/en/news/olympics-100-metres-winners-list-men-women-gold-medals-champions} (100m) and \url{https://en.wikipedia.org/wiki/List_of_Olympic_medalists_in_alpine_skiing} (downhill).
The templates have to be adapted, depending on whether the model is asked about the men's or the women's competition.
Taken together, each datapoint consists of the competition (100m or downhill), the year of the games, the subgroup (men or women), and the type of medal (gold, silver, bronze).
For example, one datapoint is $d=($ \textit{100m}, $1968$, \textit{men}, \textit{gold} $)$.
Athletes are often called by their nicknames.
We ensure that the set of correct answer contains the nickname as well as the real name(s).
For example, the set of correct answers for the datapoint above is $A_d = \lbrace$ \textit{Jim Hines}, \textit{James Hines}, \textit{James Ray Hines}$\rbrace$.
Each year in which Summer Olympics or Winter Olympics were held generates 6 datapoints (3 types of medals, men and women).
We consider games until 2022 and remove ambiguous cases, resulting in a total of $148$ datapoint for \textsc{100m} and $117$ datapoints for \textsc{downhill}.

\subsection*{Writers}
The \textsc{writers} dataset tests for the year of birth of well-known writers.
Thus, each datapoint is a writer and the set of correct answers contains their year of birth, e.g. $d=($ \textit{Friedich Schiller} $)$ and $A_d = \lbrace 1759 \rbrace$.
We tried to generate a dataset structure such that writers are sampled equally from the languages we consider.
That is, one fifth of the data are English-language writers, one fifth are German-language writers, etc.
However, we did not ensure that all countries in which these languages are spoken are taken into account.
Lists of writers for the five languages were taken from Wikipedia:
\begin{itemize}
    \small
    \item English (American authors only): \url{https://de.wikipedia.org/wiki/Liste_amerikanischer_Schriftsteller}
    \item German: \url{https://en.wikipedia.org/wiki/List_of_German-language_authors}
    \item Italian: \url{https://en.wikipedia.org/wiki/List_of_Italian_writers}
    \item Dutch: \url{https://en.wikipedia.org/wiki/List_of_Dutch-language_writers}
    \item Swedish: \url{https://en.wikipedia.org/wiki/List_of_Swedish-language_writers}
\end{itemize}
The list of Swedish-language writers had 186 entries and was the shortest.
Therefore, we randomly sampled 186 writers from each of the lists (without replacement) and used those $186\times 5=930$ datapoints to compose the dataset.

\subsection*{Companies}
The \textsc{companies} dataset tests for the headquarters locations of different companies.
Similar to \textsc{writers}, we try to cover five different countries (US, Germany, Italy, Netherlands, Sweden), such that each of the languages we work with is the dominant language in one of them.
Each datapoint consists of a company, e.g. $d=($ \textit{Volvo AB} $)$, and the set of correct answers contains all relevant variations in the city name, e.g. $A_d = \lbrace$ \textit{Gothenburg, Göteborg, Gotemburgo, Gotenburg}$\rbrace$.
We took the 100 largest companies for each of these countries from different lists on the internet:
\begin{itemize}
        \small
    \item US: \url{https://en.wikipedia.org/wiki/List_of_largest_companies_in_the_United_States_by_revenue}
    \item Germany: \url{https://de.wikipedia.org/wiki/Liste_der_gr%C3%B6%C3%9Ften_Unternehmen_in_Deutschland_(Wertsch%C3%B6pfung)}
    \item Italy: \url{https://www.value.today/headquarters/italy}
    \item Netherlands: \url{https://www.value.today/headquarters/netherlands}
    \item Sweden: \url{https://www.stockblogs.se/sveriges-storsta-foretag/}
\end{itemize}
If possible, we extracted both company and headquarters location from these lists.
When no location was given, we searched for it online.
In total, the dataset contains $100\times 5$ datapoints.

\section{Sense generation prompts}\label{app:B_sense_creation}

\subsection*{Simple facts}
For all \textsc{simple facts} datasets, except \textsc{arithmetics}, only the task instructions (corresponding to the templates in \cref{tab:templates}) need to be translated, since the input data does not change between languages.
The prompt for translating is ``Please translate the following text into [LANGUAGE]:\textbackslash n[TEXT]''.
The prompt for paraphrasing is ``Please paraphrase the following text:\textbackslash n[TEXT]''.
The \textsc{arithmetics} input data consists of spelled-out numbers, which have to be translated as well.
In the case of paraphrasing, these spelled-out numbers are not paraphrased but remain in their original version.
In the case of translation, the model is instructed to translate each number separately using the translation prompt above.

\subsection*{Benchmark data}
We use the model to generate alternative senses, treating the task instruction and the input data separately.
The prompt for translating is ``Please translate the following text into [LANGUAGE]:\textbackslash n[TEXT]''.
[LANGUAGE] is replaced by the target language and [TEXT] by the instruction (for translating instructions) or each datapoint from the benchmark (for translating input data).
For Belebele, it was necessary to explicitly instruct the model to translate everything \textit{without answering the question}.
The prompt for paraphrasing differs depending on whether task instructions or input data are paraphrased.
The prompt for paraphrasing the task instruction is ``Please paraphrase the following text:\textbackslash n[TEXT]''.
The prompt for paraphrasing the input data from the benchmarks is task-specific to help preserve the structure of the original task prompt:

\begin{itemize}
    \item \small PAWS: ``Please paraphrase the following two sentences (separately). Reply only with the paraphrased text and do not add any additional comments: \textbackslash n[TEXT].''
    \item \small XNLI: ``Please paraphrase the following premise and hypothesis (separately). Reply only with the paraphrased text and do not add any additional comments: \textbackslash n[TEXT].''
    \item \small COPA: Please paraphrase the following premise and two alternatives (separately). Reply only with the paraphrased text and do not add any additional comments: \textbackslash n[TEXT].''
    \item \small Belebele: ``Please paraphrase the following text passage, question, and multiple-choice answer options (separately). Make sure to paraphrase everything, including the passage and reply only with the paraphrased text and do not add any additional comments:\textbackslash n[TEXT].''
\end{itemize}

\section{Task instructions and alternative senses}\label{app:C_instructions}

\subsection*{Simple facts}

\cref{tab:instructions_simple_facts} shows the original English ($en$) task instructions for the \textsc{simple facts} datasets as well as the model's paraphrases ($en^P$) and translations ($de^T$, $it^T$, $nl^T$, $sv^T$) thereof.
Native speakers of the corresponding languages judged the paraphrases and translations to be mostly accurate, although they tend to stay very close to the English original.
In some cases, this tendency leads to some formal mistakes.
For example, the Dutch instruction for \textsc{arithmetics} is ``Wat is [NUMBER1] plus [NUMBER2]? Antwoord alstublieft alleen met het juiste nummer [...]''), where ``Hoeveel is [NUMBER1] plus [NUMBER2]? Antwoord alstublieft alleen met het juiste getal [...]'' would be more correct.
In addition, there is a grammatical mistake in the Swedish translation for \textsc{elements}, where the definitive article of ``the atomic number'' should be expressed by a suffix on the noun ``atomnummer'', resulting in ``atomnumret''.

\begin{longtable}{@{\extracolsep{5pt}}l|l|p{9cm} }

\caption{\textsc{Simple facts} task instructions.}
\label{tab:instructions_simple_facts} \\

\footnotesize \textbf{Task} & \footnotesize \textbf{Language} & \footnotesize \textbf{Instruction} \\

\hline
\multirow{2}{*}{{\small \textsc{arithmetics}}}
&
\footnotesize $en$ &
\footnotesize What is [NUMBER1] plus [NUMBER2]? Please reply with only the correct number (in numerical form) and no additional words.
\\
\cline{2-3}
&\footnotesize  $en^P$ &
\footnotesize What is the sum of [NUMBER1] and [NUMBER2]? Please respond with only the correct numerical answer and no extra words.
\\
\cline{2-3}
&\footnotesize  $de^T$ &
\footnotesize Was ist [NUMBER1] plus [NUMBER2]? Bitte antworten Sie nur mit der korrekten Zahl (in numerischer Form) und ohne zusätzliche Wörter.
\\
\cline{2-3}
&\footnotesize  $it^T$ &
\footnotesize Quanto fa [NUMBER1] più [NUMBER2]? Si prega di rispondere solo con il numero corretto (in forma numerica) e senza parole aggiuntive.
\\
\cline{2-3}
&\footnotesize  $nl^T$ &
\footnotesize Wat is [NUMBER1] plus [NUMBER2]? Antwoord alstublieft alleen met het juiste nummer (in numerieke vorm) en geen extra woorden.
\\
\cline{2-3}
&\footnotesize  $sv^T$ &
\footnotesize Vad är [NUMBER1] plus [NUMBER2]? Vänligen svara endast med det korrekta numret (i numerisk form) och inga ytterligare ord.
\\
\hline

\multirow{3}{1.5cm}{{\small \textsc{elements-from-element}}}
&
\footnotesize $en$ &
\footnotesize What is the atomic number of the chemical element [ELEMENT]? Please reply with the number only and do not use any additional words.
\\
\cline{2-3}
&\footnotesize  $en^P$ &
\footnotesize Please provide the atomic number of the chemical element [ELEMENT] using only the number and no extra words.
\\
\cline{2-3}
& \footnotesize $de^T$ &
\footnotesize Was ist die Ordnungszahl des chemischen Elements [ELEMENT]? Bitte antworten Sie nur mit der Zahl und verwenden Sie keine zusätzlichen Wörter.
\\
\cline{2-3}
& \footnotesize $it^T$ &
\footnotesize Qual è il numero atomico dell'elemento chimico [ELEMENT]? Si prega di rispondere solo con il numero e di non utilizzare altre parole aggiuntive.
\\
\cline{2-3}
&\footnotesize  $nl^T$ &
    \footnotesize Wat is het atoomnummer van het chemisch element [ELEMENT]? Antwoord alstublieft alleen met het nummer en gebruik geen extra woorden.
\\
\cline{2-3}
&\footnotesize  $sv^T$ &
\footnotesize Vad är det atomnummer för grundämnet [ELEMENT]? Vänligen svara endast med numret och använd inga ytterligare ord.
\\
\hline

\multirow{3}{1.5cm}{{\small \textsc{elements-from-position}}}
&
\footnotesize $en$ &
\footnotesize What is the atomic number of the chemical element in period [PERIOD] and group [GROUP]? Please reply with the number only and do not use any additional words.
\\
\cline{2-3}
&\footnotesize  $en^P$ &
\footnotesize Please provide the atomic number of the element in period [PERIOD] and group [GROUP]. Respond with only the number and no extra words.
\\
\cline{2-3}
&\footnotesize  $de^T$ &
\footnotesize "Was ist die Ordnungszahl des chemischen Elements in Periode [PERIOD] und Gruppe [GROUP]? Bitte antworten Sie nur mit der Zahl und verwenden Sie keine zusätzlichen Wörter.
\\
\cline{2-3}
&\footnotesize  $it^T$ &
\footnotesize Qual è il numero atomico dell'elemento chimico nel periodo [PERIOD] e nel gruppo [GROUP]. Si prega di rispondere solo con il numero e di non utilizzare altre parole aggiuntive.
\\
\cline{2-3}
&\footnotesize  $nl^T$ &
    \footnotesize Wat is het atoomnummer van het chemisch element in periode [PERIOD] en groep [GROUP]. Antwoord alstublieft alleen met het nummer en gebruik geen extra woorden.
\\
\cline{2-3}
&\footnotesize  $sv^T$ &
\footnotesize Vad är det atomnummer för det kemiska elementet i period [PERIOD] och grupp [GROUP]. Vänligen svara endast med numret och använd inga ytterligare ord.
\\
\hline

\multirow{2}{1.5cm}{{\small \textsc{olympics-100m}}}
&
\footnotesize $en$ &
\footnotesize Who won the [MEDAL] medal in the [GENDER] 100 meters at the [YEAR] Summer Olympics? Please reply with the name only and do not use any additional words.
\\
\cline{2-3}
&\footnotesize  $en^P$ &
\footnotesize Please provide the name of the athlete who won the [MEDAL] medal in the [GENDER] 100 meters at the [YEAR] Summer Olympics, using only the name and no extra words.
\\
\cline{2-3}
& \footnotesize $de^T$ &
\footnotesize Wer hat die [MEDAL]-Medaille im [GENDER]-100-Meter-Lauf bei den Olympischen Sommerspielen [YEAR] gewonnen? Bitte antworten Sie nur mit dem Namen und verwenden Sie keine zusätzlichen Wörter.
\\
\cline{2-3}
& \footnotesize $it^T$ &
\footnotesize Chi ha vinto la medaglia [MEDAL] nei 100 metri [GENDER] alle Olimpiadi estive del [YEAR]? Si prega di rispondere solo con il nome e di non utilizzare altre parole aggiuntive.
\\
\cline{2-3}
&\footnotesize  $nl^T$ &
    \footnotesize Wie heeft de [MEDAL] medaille gewonnen op de 100 meter voor [GENDER] tijdens de Zomerspelen van [YEAR]? Antwoord alstublieft alleen met de naam en gebruik geen extra woorden.
\\
\cline{2-3}
&\footnotesize  $sv^T$ &
\footnotesize Vem vann [MEDAL] medaljen i [GENDER] 100 meter vid sommar-OS [YEAR]? Vänligen svara med namnet endast och använd inga ytterligare ord.
\\
\hline

\multirow{2}{1.5cm}{{\small \textsc{olympics-downhill}}}
&
\footnotesize $en$ &
\footnotesize Who won the [MEDAL] medal in the [GENDER] downhill competition at the [YEAR] Winter Olympics? Please reply with the name only and do not use any additional words.
\\
\cline{2-3}
&\footnotesize  $en^P$ &
\footnotesize Please provide the name of the athlete who won the [MEDAL] medal in the [GENDER] downhill competition at the [YEAR] Winter Olympics, without using any extra words.
\\
\cline{2-3}
&\footnotesize  $de^T$ &
\footnotesize Wer hat die [MEDAL]-Medaille im [GENDER]-Abfahrtsrennen bei den Olympischen Winterspielen [YEAR] gewonnen? Bitte anworten Sie nur mit dem Namen und verwenden Sie keine zusätzlichen Wörter.
\\
\cline{2-3}
&\footnotesize  $it^T$ &
\footnotesize Chi ha vinto la medaglia [MEDAL] nella gara di discesa libera [GENDER] alle Olimpiadi invernali del [YEAR]? Per favore, rispondi solo con il nome e non utilizzare altre parole aggiuntive.
\\
\cline{2-3}
&\footnotesize  $nl^T$ &
    \footnotesize Wie heeft de [MEDAL] medaille gewonnen in de [GENDER] afdaling wedstrijd op de [YEAR] Olympische Winterspelen? Antwoord alstublieft alleen met de naam en gebruik geen extra woorden.
\\
\cline{2-3}
&\footnotesize  $sv^T$ &
\footnotesize Vem vann [MEDAL] medaljen i [GENDER] störtloppstävling vid vinter-OS [YEAR]? Vänligen svara med namnet endast och använd inga ytterligare ord.
\\
\hline

\multirow{2}{1.5cm}{{\small \textsc{writers}}}
&
\footnotesize $en$ &
\footnotesize In what year was the writer [AUTHOR] born? Please reply with the correct year only and do not use any additional words.
\\
\cline{2-3}
&\footnotesize  $en^P$ &
\footnotesize What is the birth year of the author [AUTHOR]? Please respond with only the correct year and avoid using extra words.
\\
\cline{2-3}
& \footnotesize  $de^T$ &
\footnotesize In welchem Jahr wurde der Schriftsteller / die Schriftstellerin [AUTHOR] geboren? Bitte antworten Sie nur mit dem korrekten Jahr und verwenden Sie keine zusätzlichen Wörter.
\\
\cline{2-3}
&\footnotesize  $it^T$ &
\footnotesize In che anno è nato lo scrittore / è nata la scrittrice [AUTHOR]? Per favore, rispondi solo con l'anno corretto e non utilizzare altre parole aggiuntive.
\\
\cline{2-3}
& \footnotesize $nl^T$ &
\footnotesize In welk jaar is de schrijver [AUTHOR] geboren? Uw antwoord moet alleen bestaan uit het juiste jaartal.
\\
\cline{2-3}
&\footnotesize  $sv^T$ &
\footnotesize I vilket år föddes författaren [AUTHOR]? Ditt svar ska bara bestå av det korrekta året.
\\
\hline

\multirow{2}{1.5cm}{{\small \textsc{companies}}}
&
\footnotesize $en$ &
\footnotesize In what city does [COMPANY] have its headquarters? Please reply only with the name of the city and no additional words.
\\
\cline{2-3}
& \footnotesize $en^P$ &
\footnotesize Where is the headquarters of [COMPANY] located? Please respond with only the city name, without any extra words.
\\
\cline{2-3}
&\footnotesize  $de^T$ &
\footnotesize In welcher Stadt hat [COMPANY] seinen Hauptsitz? Bitte antworten Sie nur mit dem Namen der Stadt und ohne zusätzliche Wörter.
\\
\cline{2-3}
& \footnotesize $it^T$ &
\footnotesize In quale città ha sede [COMPANY]? Si prega di rispondere solo con il nome della città e senza parole aggiuntive.
\\
\cline{2-3}
& \footnotesize $nl^T$ &
\footnotesize In welke stad heeft [COMPANY] zijn hoofdkantoor? Antwoord alstublieft alleen met de naam van de stad en geen extra woorden.
\\
\cline{2-3}
&\footnotesize  $sv^T$ &
\footnotesize I vilken stad har [COMPANY] sitt huvudkontor? Vänligen svara endast med stadens namn och inga ytterligare ord.
\\
\hline

\end{longtable}

\subsection*{Benchmark data}

\cref{tab:instructions_benchmarks} lists the original English ($en$) task instructions for the benchmark datasets as well as the model's paraphrases ($en^P$) and translations ($de^T$, $it^T$, $nl^T$, $sv^T$) thereof.
Native speakers of the corresponding languages judged the paraphrases and translations to be generally accurate but some sentences contained minor mistakes or aspects that the native speakers would have translated differently.
Points that were mentioned are that
1) the model translates ``premise'' to ``presupposto'' in Italian (COPA and XNLI) even though ``premessa'' is more appropriate and
2) the repeated use of ``noch'' in the Dutch XNLI instruction is incorrect and the correct sentence should end with something like ``als de premisse de hypothese noch impliceert nog tegenspreekt''.

\begin{longtable}{@{\extracolsep{5pt}}l|l|p{9cm} }
\caption{Benchmark data task instructions.}
\label{tab:instructions_benchmarks} \\
\footnotesize \textbf{Task} & \footnotesize \textbf{Language} & \footnotesize \textbf{Instruction} \\
\hline

\multirow{2}{1.5cm}{{\textsc{paws}}}
&
\footnotesize $en$ &
\footnotesize Do the following two sentences have the same meaning?\newline Sentence 1: ``[SENTENCE1]''\newline Sentence 2: ``[SENTENCE2]''\newline Please reply with a single word, either ``yes'' or ``no''.
\\
\cline{2-3}
& \footnotesize $en^P$ &
\footnotesize Are the meanings of the following two sentences the same?\newline Sentence 1: ``[SENTENCE1]''\newline Sentence 2: ``[SENTENCE2]''\newline Please respond with either ``yes'' or ``no''.
\\
\cline{2-3}
& \footnotesize $de^T$ &
\footnotesize Haben die folgenden beiden Sätze die gleiche Bedeutung?\newline Satz 1: ``[SENTENCE1]''\newline Satz 2: ``[SENTENCE2]''\newline Bitte antworten Sie mit einem einzigen Wort, entweder ``ja'' oder ``nein''.
\\
\cline{2-3}
& \footnotesize $it^T$ &
\footnotesize Le seguenti due frasi hanno lo stesso significato?\newline Frase 1: ``[SENTENCE1]''\newline Frase 2: ``[SENTENCE2]''\newline Rispondi con una sola parola, ``sì'' o ``no''.
\\
\cline{2-3}
& \footnotesize $nl^T$ &
\footnotesize Hebben de volgende twee zinnen dezelfde betekenis?\newline Zin 1: ``[SENTENCE1]''\newline Zin 2: ``[SENTENCE2]''\newline Antwoord alstublieft met één woord, ofwel ``ja'' ofwel ``nee''.
\\
\cline{2-3}
& \footnotesize $sv^T$ &
\footnotesize Har de följande två meningarna samma betydelse?\newline Mening 1: ``[SENTENCE1]''\newline Mening 2: ``[SENTENCE2]''\newline Svara med ett enda ord, antingen ``ja'' eller ``nej''.
\\
\hline

\multirow{2}{1.5cm}{{\textsc{xnli}}}
&
\footnotesize $en$ &
\footnotesize Given the following premise and hypothesis, please identify whether the premise entails the hypothesis, contradicts the hypothesis, or neither of the two.\newline Premise: ``[PREMISE]''\newline Hypothesis: ``[HYPOTHESIS]''\newline Please reply with a single word: ``entailment'' if the premise entails the hypothesis, ``contradiction'' if the premise contradicts the hypothesis, and ``neutral'' if the premise neither entails nor contradicts the hypothesis.
\\
\cline{2-3}
& \footnotesize $en^P$ &
\footnotesize Please determine if the premise and hypothesis are related.\newline Premise: ``[PREMISE]''\newline Hypothesis: ``[HYPOTHESIS]''
\newline If the premise supports the hypothesis, indicate ``entailment''. If the premise contradicts the
hypothesis, indicate ``contradiction''. If there is no clear relationship between the two, indicate ``neutral''.
\\
\cline{2-3}
& \footnotesize $de^T$ &
\footnotesize Angesichts der folgenden Prämisse und Hypothese, bitte identifizieren Sie, ob die Prämisse die Hypothese impliziert,
der Hypothese widerspricht oder weder das eine noch das andere.\newline Prämisse: ``[PREMISE]''\newline 
Hypothese: ``[HYPOTHESIS]''\newline Bitte antworten Sie mit einem einzigen Wort: ``Implikation'', wenn die
Prämisse die Hypothese impliziert, ``Widerspruch'', wenn die Prämisse der Hypothese widerspricht, und ``neutral'',
wenn die Prämisse weder die Hypothese impliziert noch ihr widerspricht.
\\
\cline{2-3}
& \footnotesize $it^T$ &
\footnotesize Dato il seguente presupposto e ipotesi, per favore identifica se il presupposto implica l'ipotesi, contraddice
l'ipotesi o né implica né contraddice l'ipotesi.\newline Presupposto: ``[PREMISE]''\newline Ipotesi:
``[HYPOTHESIS]''\newline Per favore rispondi con una sola parola: ``implicazione'' se il presupposto
implica l'ipotesi, ``contraddizione'' se il presupposto contraddice l'ipotesi e ``neutrale'' se il presupposto né
implica né contraddice l'ipotesi.
\\
\cline{2-3}
& \footnotesize $nl^T$ &
\footnotesize Gegeven de volgende premisse en hypothese, identificeer alstublieft of de premisse de hypothese impliceert, de
hypothese tegenspreekt, of geen van beide.\newline Premisse: ``[PREMISE]''\newline Hypothesis:
``[HYPOTHESIS]''\newline Antwoord alstublieft met één woord: ``implicatie'' als de premisse de hypothese
impliceert, ``tegenspraak'' als de premisse de hypothese tegenspreekt, en ``neutraal'' als de premisse noch de
hypothese impliceert noch tegenspreekt.
\\
\cline{2-3}
& \footnotesize $sv^T$ &
\footnotesize Givet följande premiss och hypotes, vänligen ange om premissen innebär hypotesen, motsäger hypotesen eller varken
innebär eller motsäger hypotesen.\newline Premiss: ``[PREMISE]''\newline Hypotes: ``[HYPOTHESIS]''
\newline Vänligen svara med ett enda ord: ``innebär'' om premissen innebär hypotesen, ``motsäger'' om
premissen motsäger hypotesen och ``neutral'' om premissen varken innebär eller motsäger hypotesen.
\\
\hline

\multirow{2}{1.5cm}{{\textsc{copa}}}
&
\footnotesize $en$ &
\footnotesize Given the following premise, which of the two alternatives is more plausible?\newline Premise: ``[PREMISE]''\newline Alternative 1: ``[CHOICE1]''\newline Alternative 2: ``[CHOICE2]''\newline Please answer with a single word: ``Alternative-1'' if alternative 1 is more plausible and ``Alternative-2'' if alternative 2 is more plausible.
\\
\cline{2-3}
& \footnotesize $en^P$ &
\footnotesize Based on the provided premise, which of the two options is more likely?\newline Premise: ``[PREMISE]''\newline Option 1: ``[CHOICE1]''\newline Option 2: ``[CHOICE2]''\newline Please respond with either ``Option-1'' if option 1 is more likely, or ``Option-2'' if option 2 is more likely.
\\
\cline{2-3}
& \footnotesize $de^T$ &
\footnotesize Angesichts der folgenden Prämisse, welche der beiden Alternativen ist plausibler?\newline Prämisse: ``[PREMISE]''\newline Alternative 1: ``[CHOICE1]''\newline Alternative 2: ``[CHOICE2]''\newline Bitte antworten Sie mit einem einzigen Wort: ``Alternative-1'', wenn Alternative 1 plausibler ist, und ``Alternative-2'', wenn Alternative 2 plausibler ist.
\\
\cline{2-3}
& \footnotesize $it^T$ &
\footnotesize Dato il seguente presupposto, quale delle due alternative è più plausibile?\newline Presupposto: ``[PREMISE]''\newline Alternativa 1: ``[CHOICE1]''\newline Alternativa 2: ``[CHOICE2]''\newline Per favore, rispondi con una sola parola: ``Alternativa-1'' se l'alternativa 1 è più plausibile e ``Alternativa-2'' se l'alternativa 2 è più plausibile.
\\
\cline{2-3}
& \footnotesize $nl^T$ &
\footnotesize Gegeven de volgende premisse, welke van de twee alternatieven is waarschijnlijker?\newline Premisse: ``[PREMISE]''\newline Alternatief 1: ``[CHOICE1]''\newline Alternatief 2: ``[CHOICE2]''\newline Antwoord alstublieft met één woord: ``Alternatief-1'' als alternatief 1 waarschijnlijker is en ``Alternatief-2'' als alternatief 2 waarschijnlijker is.
\\
\cline{2-3}
& \footnotesize $sv^T$ &
\footnotesize Givet följande premiss, vilket av de två alternativen är mer troligt?\newline Premiss: ``[PREMISE]''\newline Alternativ 1: ``[CHOICE1]''\newline Alternativ 2: ``[CHOICE2]''\newline Svara med ett enda ord: ``Alternativ-1'' om alternativ 1 är mer troligt och ``Alternativ-2'' om alternativ 2 är mer troligt.
\\
\hline

\multirow{2}{1.5cm}{{\textsc{belebele}}}
&
\footnotesize $en$ &
\footnotesize [PASSAGE]\newline \newline [QUESTION]\newline \newline Option A: [ANSWER1]\newline Option B: [ANSWER2]\newline Option C: [ANSWER3]\newline Option D: [ANSWER4]\newline \newline Please reply with ``A'', ``B'', ``C'', or ``D'' to indicate the correct answer. Your reply should be a single letter and should not contain any additional words.
\\
\cline{2-3}
& \footnotesize $en^P$ &
\footnotesize [PASSAGE]\newline \newline [QUESTION]\newline \newline A) [ANSWER1]\newline B) [ANSWER2]\newline C) [ANSWER3]\newline D) [ANSWER4]\newline \newline Please respond with the letter corresponding to the correct answer choice. Your response should be a single letter and should not include any extra words.
\\
\cline{2-3}
& \footnotesize $de^T$ &
\footnotesize [PASSAGE]\newline \newline [QUESTION]\newline \newline Option A: [ANSWER1]\newline Option B: [ANSWER2]\newline Option C: [ANSWER3]\newline Option D: [ANSWER4]\newline \newline Antworten Sie bitte mit ``A'', ``B'', ``C'', oder ``D'', um die richtige Antwort anzugeben. Ihre Antwort sollte nur ein einzelner Buchstabe sein und keine zusätzlichen Wörter enthalten.
\\
\cline{2-3}
& \footnotesize $it^T$ &
\footnotesize PASSAGE]\newline \newline [QUESTION]\newline \newline Opzione A: [ANSWER1]\newline Opzione B: [ANSWER2]\newline Opzione C: [ANSWER3]\newline Opzione D: [ANSWER4]\newline \newline Rispondi con ``A'', ``B'', ``C'' o ``D'' per indicare la risposta corretta. La tua risposta deve essere una singola lettera e non deve contenere parole aggiuntive.
\\
\cline{2-3}
& \footnotesize $nl^T$ &
\footnotesize [PASSAGE]\newline \newline [QUESTION]\newline \newline Optie A: [ANSWER1]\newline Optie B: [ANSWER2]\newline Optie C: [ANSWER3]\newline Optie D: [ANSWER4]\newline \newline Antwoord alstublieft met ``A'', ``B'', ``C'' of ``D'' om het juiste antwoord aan te geven. Uw antwoord moet uit één letter bestaan en mag geen extra woorden bevatten.
\\
\cline{2-3}
& \footnotesize $sv^T$ &
\footnotesize [PASSAGE]\newline \newline [QUESTION]\newline \newline Alternativ A: [ANSWER1]\newline Alternativ B: [ANSWER2]\newline Alternativ C: [ANSWER3]\newline Alternativ D: [ANSWER4]\newline \newline Vänligen svara med ``A'', ``B'', ``C'', eller ``D'' för att ange det korrekta svaret. Ditt svar ska vara en enda bokstav och får inte innehålla några ytterligare ord.
\\
\hline


\end{longtable}

\section{Accuracy scores}\label{app:accuracy_tables}

\subsection*{Simple facts}

\begin{table}[H]
\caption{Accuracy (\%) on the simple fact datasets, with 95\% confidence intervals.}\label{tab:simple-facts-accuracy}
\centering
\footnotesize
\begin{tabular}{l|c|cc|cc|c|c||c}
\toprule
{} &  \footnotesize arithmetics  & \multicolumn{2}{c|}{\footnotesize elements} & \multicolumn{2}{c|}{\footnotesize olympics} & \footnotesize writers & \footnotesize companies & \footnotesize total\\
\midrule
{} & \footnotesize - &  \footnotesize elem & \footnotesize pos & \footnotesize 100m & \footnotesize downhill &  - &  - & \footnotesize avg \\
\midrule
en      &   99.4$^{\pm 1.0}$    &  100.0$^{\pm nan}$   &    37.8$^{\pm 10.0}$    &   55.4$^{\pm 8.1}$    &   37.6$^{\pm 9.4}$    &   76.2$^{\pm 2.9}$     &   78.2$^{\pm 3.8}$     &  73.5  \\
en$^P$  &   98.6$^{\pm 1.4}$     &  100.0$^{\pm nan}$   &    42.2$^{\pm 10.0}$    &   54.1$^{\pm 8.1}$    &   31.6$^{\pm 9.4}$    &   76.2$^{\pm 2.8}$     &   76.0$^{\pm 3.8}$     &  72.2  \\
de$^T$  &   45.2$^{\pm 4.6}$     &  98.9$^{\pm 5.6}$    &    40.0$^{\pm 10.0}$    &   50.7$^{\pm 8.1}$    &   35.0$^{\pm 9.4}$    &   76.8$^{\pm 2.8}$     &   75.4$^{\pm 3.8}$     &  68.7  \\
it$^T$  &   44.0$^{\pm 4.4}$     &  100.0$^{\pm nan}$   &    36.7$^{\pm 10.0}$    &   51.4$^{\pm 8.1}$    &   35.0$^{\pm 9.4}$    &   75.3$^{\pm 2.9}$     &   73.6$^{\pm 4.0}$     &  67.4  \\
nl$^T$  &   42.4$^{\pm 4.4}$     &  100.0$^{\pm nan}$   &    36.7$^{\pm 10.0}$    &   52.0$^{\pm 8.1}$    &   35.0$^{\pm 8.5}$    &   76.8$^{\pm 2.9}$     &   73.2$^{\pm 4.2}$     &  67.7  \\
sv$^T$  &   19.2$^{\pm 3.6}$     &  100.0$^{\pm nan}$   &    41.1$^{\pm 11.1}$    &   50.7$^{\pm 8.1}$    &   33.3$^{\pm 8.5}$    &   74.3$^{\pm 2.9}$     &   71.8$^{\pm 4.0}$     &  65.0  \\
\bottomrule
\end{tabular}
\end{table}

\subsection*{Benchmark data}

\begin{table}[H]
\caption{Accuracy (\%) on the benchmark datasets, with 95\% confidence intervals.}\label{tab:benchmarks-accuracy}
\centering
\begin{tabular}{l|c|c|c|c||c}
\toprule
{}       &   paws  &   xnli   &   copa    &   belebele  & avg   \\
\midrule
en      &   75.6$^{\pm 1.9}$   &   43.7$^{\pm 1.4}$   &   84.4$^{\pm 3.4}$    &   85.9$^{\pm 2.3}$      &  72.4     \\
en$^P$  &   67.6$^{\pm 2.1}$   &   53.5$^{\pm 1.4}$   &   82.2$^{\pm 3.4}$    &   -         &  -        \\
de$^T$  &   64.3$^{\pm 2.1}$   &   50.0$^{\pm 1.4}$   &   85.6$^{\pm 3.2}$    &   81.2$^{\pm 2.7}$      &  70.3     \\
it$^T$  &   75.1$^{\pm 2.0}$   &   56.4$^{\pm 1.4}$   &   86.6$^{\pm 3.2}$    &   81.0$^{\pm 2.7}$      &  74.8     \\
nl$^T$  &   71.9$^{\pm 2.0}$   &   50.9$^{\pm 1.4}$  &   83.4$^{\pm 3.4}$    &   79.0$^{\pm 2.7}$      &  71.3     \\
sv$^T$  &   55.9$^{\pm 2.2}$   &   47.0$^{\pm 1.4}$   &   89.2$^{\pm 2.8}$    &   79.1$^{\pm 2.8}$      &  67.8     \\
\bottomrule
\end{tabular}
\end{table}

\section{Accuracy based on containment versus exact match}\label{app:containment_versus_exact_match}

On the \textsc{simple facts} datasets, the model is instructed to reply with the correct entity (and no additional words), which we then use to quantify consistency.
Hence, it is important that the model actually follows that instruction across all senses.
Otherwise, it could be that the model replies with ``Friedrich Schiller was born in 1759'' when prompted for a writer in English but ``1759'' when prompted in German.
While a failure to follow the instruction in one language but not the other could be considered an unwanted inconsistency, the meaning of both answers is arguably the same, and we would like to differentiate between both cases.

\captionsetup{font=footnotesize}
 \begin{wrapfigure}{r}{0.3\textwidth}
   \centering
     \includegraphics[width=0.3\textwidth]{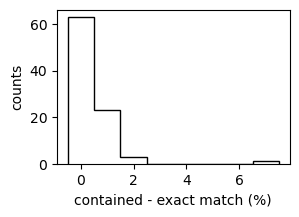}
   \caption{Containment score minus exact match score across tasks and senses.}\label{fig:contain-em}
\end{wrapfigure}
\captionsetup{font=normalfont}
If the model replies correctly but not in one word, the response \textit{contains} the right answer but does not \textit{exactly match} it.
\cref{fig:contain-em} shows the distribution of the difference in accuracy based on containment versus exact match.
The scores for \textsc{companies} and \textsc{writers} are calculated separately for each language-specific subgroup of samples (i.e.~US companies, German companies, ...) to obtain more detailed information.
In most cases, the ``containment'' score is not at all or only slightly higher than the exact match score.
The only exception occurs for Dutch companies when prompted with $en^P$, with a 7\% difference in accuracy.
This mismatch arises because the model -- while otherwise replying with only the city name -- always responds with a full sentence when the correct answer is ``The Hague'' (e.g.\ ``The headquarters of Shell PLC is located in The Hague.'').
Thus, except for this curious case, inconsistencies can largely not be attributed to a failure to express a response in the correct form.

\section{Consistency scores}\label{app:consistency_tables}

\subsection*{Simple facts}

\begin{table}[H]
\caption{Consistency (\%) on the simple fact datasets.}\label{tab:simple-facts-consistency}
\centering
\begin{tabular}{l|c|cc|cc|c|c||c}
\toprule
{} & \small arithmetics  & \multicolumn{2}{c|}{\small elements} & \multicolumn{2}{c|}{\small olympics} & \small writers &  \small companies & \small total\\
\midrule
{} & \small -  &  \small elem & \small position & \small 100m & \small downhill &  - &  - & \small avg \\
\midrule
id      &   100.0 &   100.0   &    91.1   &   90.5    &   88.9    &   87.1    &   97.4    &  92.8 \\
en$^P$  &   99.0  &   100.0   &    77.8   &   82.4    &   61.5    &   82.7    &   91.2    &  86.0 \\
de$^T$  &   45.0  &   98.9    &    71.1   &   80.4    &   76.9    &   81.2    &   89.4    &  81.7 \\
it$^T$  &   44.2  &   100.0   &    70.0   &   75.7    &   70.1    &   81.2    &   86.4    &  79.8 \\
nl$^T$  &   42.2  &   100.0   &    67.8   &   83.8    &   64.1    &   79.1    &   88.8    &  79.8 \\
sv$^T$  &   19.4  &   100.0   &    70.0   &   83.8    &   59.8    &   77.3    &   84.8    &  76.2 \\
\bottomrule
\end{tabular}
\end{table}

\subsection*{Benchmark data}

\begin{table}[H]
\caption{Consistency (\%) on the benchmark datasets.}\label{tab:benchmarks-consistency}
\centering
\begin{tabular}{l|c|c|c|c||c}
\toprule
{}          &   paws    &   xnli    &   copa    &   belebele   & avg    \\
\midrule
id          &   95.2    &   96.0    &   96.8    &   97.1    &   96.3    \\
en$^P$      &   76.5    &   56.3    &   85.2    &   -       &   -       \\
de$^T$      &   74.7    &   51.2    &   88.8    &   84.7    &   74.8    \\
it$^T$      &   82.2    &   57.5    &   85.8    &   85.1    &   77.6    \\
nl$^T$      &   82.4    &   73.1    &   91.0    &   85.8    &   83.1    \\
sv$^T$      &   67.9    &   82.8    &   86.0    &   83.3    &   80.0    \\
\bottomrule
\end{tabular}
\end{table}

\section{Examples of inconsistent responses}\label{appendix:examples}

\begin{longtable}{@{\extracolsep{5pt}}l|l|p{9cm}}

\caption{Examples of inconsistencies for \textsc{simple facts}. We report the first ten inconsistent samples per dataset and sense.}
\label{tab:inconsistency_examples} \\

\small \textbf{Task} & \small \textbf{Senses} & \small \textbf{Examples} \\

\hline
\multirow{2}{1.9cm}{{\small \textsc{arithmetics}}} & \small ($en$\space\textbar\space $en$) & \small - \\
\cline{2-3}
                            & \small ($en$\space\textbar\space $en^P$) & \footnotesize (540\space\textbar\space 340), (1770\space\textbar\space 1778), (237\space\textbar\space Two hundred thirty-seven.), (1173\space\textbar\space One thousand one hundred seventy-three), (1013\space\textbar\space One thousand thirteen.) \\
\cline{2-3}
                            & \small ($en$\space\textbar\space $de^T$) & \footnotesize (778\space\textbar\space 678), (618\space\textbar\space 1008), (926\space\textbar\space 526), (115\space\textbar\space 915), (924\space\textbar\space 524), (1535\space\textbar\space 1035), (1693\space\textbar\space 1689), (1437\space\textbar\space 1337), (1151\space\textbar\space 1248), (1248\space\textbar\space 1448) \\
\cline{2-3}
                            & \small ($en$\space\textbar\space $it^T$) & \footnotesize (778\space\textbar\space 678), (858\space\textbar\space 788), (1471\space\textbar\space 1437), (926\space\textbar\space 836), (924\space\textbar\space 923), (577\space\textbar\space 577 + 300 = 877), (1535\space\textbar\space 1335), (1693\space\textbar\space 1683), (1437\space\textbar\space 1497), (1151\space\textbar\space 1051) \\
\cline{2-3}
                            & \small ($en$\space\textbar\space $nl^T$) & \footnotesize (778\space\textbar\space 678), (965\space\textbar\space 865), (858\space\textbar\space 958), (926\space\textbar\space 726), (115\space\textbar\space 109), (1535\space\textbar\space 935), (1693\space\textbar\space 1689), (1437\space\textbar\space 1338), (1151\space\textbar\space 846), (1248\space\textbar\space 848) \\
\cline{2-3}
                            & \small ($en$\space\textbar\space $sv^T$) & \footnotesize (778\space\textbar\space 784 + 94 = 878), (965\space\textbar\space 929), (858\space\textbar\space 792), (1471\space\textbar\space 1465), (926\space\textbar\space 733 + 163 = 896), (1277\space\textbar\space 1170), (1304\space\textbar\space 645), (924\space\textbar\space 923), (577\space\textbar\space 577 + 300 = 877), (1535\space\textbar\space 825) \\
\midrule
\multirow{1}{1.9cm}{{\small \textsc{elements-}}}    & \small ($en$\space\textbar\space $en$)   & \small - \\
\cline{2-3}
\multirow{1}{1.9cm}{{\small \textsc{from-}}}        & \small ($en$\space\textbar\space $en^P$) & \small - \\
\cline{2-3}
\multirow{1}{1.9cm}{{\small \textsc{element}}}      & \small ($en$\space\textbar\space $de^T$) & \footnotesize (114\space\textbar\space 9)  \\
\cline{2-3}
                            & \small ($en$\space\textbar\space $it^T$) & \small - \\
\cline{2-3}
                            & \small ($en$\space\textbar\space $nl^T$) & \small - \\
\cline{2-3}
                            & \small ($en$\space\textbar\space $sv^T$) & \small -  \\
\midrule
\multirow{3}{1.9cm}{{\small \textsc{elements-from-position}}} & \small ($en$\space\textbar\space $en$)   & \footnotesize (22\space\textbar\space 20), (13\space\textbar\space 31), (17\space\textbar\space 107), (107\space\textbar\space 104), (106\space\textbar\space 46), (86\space\textbar\space 14), (33\space\textbar\space 51), (17\space\textbar\space 53) \\
\cline{2-3}
                            & \small ($en$\space\textbar\space $en^P$) & \footnotesize (16\space\textbar\space 8), (19\space\textbar\space 37), (36\space\textbar\space 26), (28\space\textbar\space 39), (13\space\textbar\space 31), (23\space\textbar\space 55), (22\space\textbar\space 38), (45\space\textbar\space 46), (33\space\textbar\space 51), (16\space\textbar\space 34) \\
\cline{2-3}
                            & \small ($en$\space\textbar\space $de^T$) & \footnotesize (12\space\textbar\space 4), (16\space\textbar\space 8), (19\space\textbar\space 11), (35\space\textbar\space 17), (36\space\textbar\space 26), (28\space\textbar\space 39), (48\space\textbar\space 30), (33\space\textbar\space 15), (38\space\textbar\space 12), (22\space\textbar\space 23)  \\
\cline{2-3}
                            & \small ($en$\space\textbar\space $it^T$) & \footnotesize (2\space\textbar\space 1), (12\space\textbar\space 4), (13\space\textbar\space 5), (7\space\textbar\space 15), (16\space\textbar\space 8), (35\space\textbar\space 23), (28\space\textbar\space 35), (28\space\textbar\space 40), (48\space\textbar\space 40), (13\space\textbar\space 31) \\
\cline{2-3}
                            & \small ($en$\space\textbar\space $nl^T$) & \footnotesize (12\space\textbar\space 4), (13\space\textbar\space 5), (16\space\textbar\space 8), (21\space\textbar\space 13), (35\space\textbar\space 23), (28\space\textbar\space 39), (28\space\textbar\space 40), (48\space\textbar\space 40), (33\space\textbar\space 15), (38\space\textbar\space 12) \\
\cline{2-3}
                            & \small ($en$\space\textbar\space $sv^T$) & \footnotesize (2\space\textbar\space 1), (13\space\textbar\space 5), (16\space\textbar\space 8), (17\space\textbar\space 9), (21\space\textbar\space 23), (36\space\textbar\space 26), (28\space\textbar\space 39), (28\space\textbar\space 40), (48\space\textbar\space 40), (13\space\textbar\space 31) \\
\midrule
\multirow{2}{1.9cm}{{\small \textsc{olympics-100m}}} & \small ($en$\space\textbar\space $en$)   & \footnotesize (Charley Paddock\space\textbar\space Harold Abrahams), (Arthur Jonath\space\textbar\space Eddie Tolan), (Lloyd LaBeach\space\textbar\space Herbert McKenley), (Lloyd LaBeach\space\textbar\space Herbert McKenley), (Herb McKenley\space\textbar\space Hector Hogan), (Ben Johnson\space\textbar\space Calvin Smith), (Kim Collins\space\textbar\space Justin Gatlin), (Ethel Smith\space\textbar\space Elizabeth Robinson), (Shirley Strickland\space\textbar\space Marjorie Jackson), (Shirley Strickland\space\textbar\space Marlene Mathews) \\
\cline{2-3}
                            & \small ($en$, $en^P$) & \footnotesize (Francis Lane\space\textbar\space Frank Lane), (Fay Moulton\space\textbar\space Frank Jarvis), (Nathaniel Cartmell\space\textbar\space Reggie Walker), (Nate Cartmell\space\textbar\space Reggie Walker), (Arthur Porritt\space\textbar\space Percy Williams), (Arthur Jonath\space\textbar\space Percy Williams), (Barney Ewell\space\textbar\space Herb McKenley), (Lloyd LaBeach\space\textbar\space Herb McKenley), (Lloyd LaBeach\space\textbar\space Herb McKenley), (Herb McKenley\space\textbar\space Hector Hogan) \\
\cline{2-3}
                            & \small ($en$\space\textbar\space $de^T$) & \footnotesize (Francis Lane\space\textbar\space Frank Lane), (Nate Cartmell\space\textbar\space Reggie Walker), (Arthur Porritt\space\textbar\space Arthur Jonath), (Arthur Porritt\space\textbar\space Arthur Jonath), (Arthur Jonath\space\textbar\space Eddie Tolan), (Arthur Jonath\space\textbar\space Eddie Tolan), (Arthur Jonath\space\textbar\space Ralph Metcalfe), (Lloyd LaBeach\space\textbar\space Barney Ewell), (Lloyd LaBeach\space\textbar\space Herb McKenley), (Enrique Figuerola\space\textbar\space Bob Hayes)  \\
\cline{2-3}
                            & \small ($en$\space\textbar\space $it^T$) & \footnotesize (Fay Moulton\space\textbar\space Frank Castle), (Nathaniel Cartmell\space\textbar\space Reginald Walker), (Nate Cartmell\space\textbar\space Reginald Walker), (Charley Paddock\space\textbar\space Harold Abrahams.), (Arthur Porritt\space\textbar\space Percy Williams.), (Arthur Porritt\space\textbar\space Arthur Jonath), (Arthur Jonath\space\textbar\space Eddie Tolan.), (Arthur Jonath\space\textbar\space Eddie Tolan.), (Arthur Jonath\space\textbar\space Ralph Metcalfe.), (Lloyd LaBeach\space\textbar\space Barney Ewell.) \\
\cline{2-3}
                            & \small ($en$\space\textbar\space $nl^T$) & \footnotesize (Fay Moulton\space\textbar\space Frank Waller), (Arthur Porritt\space\textbar\space Arthur Jonath), (Arthur Porritt\space\textbar\space Arthur Jonath), (Lloyd LaBeach\space\textbar\space Barney Ewell), (Lloyd LaBeach\space\textbar\space Herb McKenley), (Herb McKenley\space\textbar\space Thane Baker.), (Enrique Figuerola\space\textbar\space Edwin Roberts.), (Valeriy Borzov\space\textbar\space Valeri Borzov), (Ben Johnson\space\textbar\space Calvin Smith.), (Ato Boldon\space\textbar\space Maurice Greene) \\
\cline{2-3}
                            & \small ($en$\space\textbar\space $sv^T$) & \footnotesize (Francis Lane\space\textbar\space Frank Lane), (Ralph Craig\space\textbar\space Donald Lippincott), (Arthur Porritt\space\textbar\space Arthur Jonath), (Lloyd LaBeach\space\textbar\space Barney Ewell.), (Lloyd LaBeach\space\textbar\space Herb McKenley), (Valeriy Borzov\space\textbar\space Valeri Borzov), (Ben Johnson\space\textbar\space Calvin Smith.), (Linford Christie\space\textbar\space Carl Lewis), (Ato Boldon\space\textbar\space Maurice Greene.), (Kim Collins\space\textbar\space Asafa Powell) \\
\midrule
\multirow{2}{1.9cm}{{\small \textsc{olympics-downhill}}} & \small ($en$\space\textbar\space $en$)   & \footnotesize (Egon Zimmermann\space\textbar\space Guy Périllat), (Franz Klammer\space\textbar\space Bernhard Russi), (Franck Piccard\space\textbar\space Franz Heinzer), (Didier Défago\space\textbar\space Didier Defago), (Beat Feuz\space\textbar\space Kjetil Jansrud), (Hedy Schlunegger\space\textbar\space Trude Beiser-Jochum), (Andrea Mead-Lawrence\space\textbar\space Trude Beiser-Jochum), (Trude Beiser-Jochum\space\textbar\space Hanni Wenzel), (Christl Haas\space\textbar\space Christine Goitschel), (Brigitte Oertli\space\textbar\space Vreni Schneider) \\
\cline{2-3}
                            & \small ($en$\space\textbar\space $en^P$) & \footnotesize (Zeno Colò\space\textbar\space Andreas Molterer), (Christian Pravda\space\textbar\space Anton Sailer), (Christian Pravda\space\textbar\space Andreas), (Guy Périllat\space\textbar\space Jean Vuarnet.), (Egon Zimmermann\space\textbar\space Jean-Claude Killy), (Bernhard\space\textbar\space Franz Klammer), (Leonhard Stock\space\textbar\space Bill Johnson), (Anton Steiner\space\textbar\space Bill Johnson), (Franck Piccard\space\textbar\space Kjetil André Aamodt), (Hans Knauss\space\textbar\space Hermann Maier.) \\
\cline{2-3}
                            & \small ($en$\space\textbar\space $de^T$) & \footnotesize (Egon Zimmermann\space\textbar\space Guy Périllat), (Bernhard\space\textbar\space Bernhard Russi), (Anton Steiner\space\textbar\space Peter Müller), (Franz\space\textbar\space Franz Heinzer), (Hermann Maier\space\textbar\space Lasse Kjus), (Hans Knauss\space\textbar\space Hermann Maier), (Antoine Dénériaz\space\textbar\space Michael Walchhofer), (Kjetil André Aamodt\space\textbar\space Michael Walchhofer), (Bode Miller\space\textbar\space Aksel Lund Svindal), (Kjetil Jansrud\space\textbar\space Aksel Lund Svindal) \\
\cline{2-3}
                            & \small ($en$\space\textbar\space $it^T$) & \footnotesize (Egon Zimmermann\space\textbar\space Jean-Claude Killy.), (Egon Zimmermann\space\textbar\space Guy Périllat.), (Bernhard\space\textbar\space Bernhard Russi.), (Leonhard Stock\space\textbar\space Peter Wirnsberger.), (Franz\space\textbar\space Franz Heinzer.), (Tommy Moe\space\textbar\space Markus Wasmeier.), (Hans Knauss\space\textbar\space Lasse Kjus.), (Fritz Strobl\space\textbar\space Lasse Kjus.), (Antoine Dénériaz\space\textbar\space Michael Walchhofer.), (Kjetil André Aamodt\space\textbar\space Michael Walchhofer.)\\
\cline{2-3}
                            & \small ($en$\space\textbar\space $nl^T$) & \footnotesize (Zeno Colò\space\textbar\space Andrea Mead-Lawrence), (Egon Zimmermann\space\textbar\space Guy Périllat), (Egon Zimmermann\space\textbar\space Guy Périllat), (Bernhard\space\textbar\space Bernhard Russi), (Leonhard Stock\space\textbar\space Peter Müller), (Leonhard Stock\space\textbar\space Peter Müller), (Franz\space\textbar\space Franz Heinzer), (Franz Heinzer\space\textbar\space Franck Piccard), (Franck Piccard\space\textbar\space Franz Heinzer), (Tommy Moe\space\textbar\space Patrick Ortlieb) \\
\cline{2-3}
                            & \small ($en$\space\textbar\space $sv^T$) & \footnotesize (Egon Zimmermann\space\textbar\space Guy Périllat), (Egon Zimmermann\space\textbar\space Guy Périllat.), (Bernhard\space\textbar\space Bernhard Russi), (Anton Steiner\space\textbar\space Bill Johnson), (Franz\space\textbar\space Franz Heinzer.), (Franck Piccard\space\textbar\space Franz Heinzer), (Tommy Moe\space\textbar\space Markus), (Hermann Maier\space\textbar\space Lasse Kjus.), (Hans Knauss\space\textbar\space Lasse Kjus.), (Antoine Dénériaz\space\textbar\space Fritz Strobl.) \\
\midrule
\multirow{2}{1.9cm}{{\small \textsc{writers}}} & \small ($en$\space\textbar\space $en$)   & \footnotesize (1978\space\textbar\space 1977), (1903\space\textbar\space 1912), (1911\space\textbar\space 1901), (1965\space\textbar\space 1968), (1982\space\textbar\space 1984), (1975\space\textbar\space 1980), (1962\space\textbar\space 1939), (1880\space\textbar\space 1891), (1930\space\textbar\space 1936), (1851\space\textbar\space 1871) \\
\cline{2-3}
                            & \small ($en$\space\textbar\space $en^P$) & \footnotesize (1952\space\textbar\space 1944), (1978\space\textbar\space 1977), (1940\space\textbar\space 1925), (1992\space\textbar\space I'm sorry, but I don't have access to personal information about individuals unless it has been shared with me in the course of our conversation.), (1903\space\textbar\space 1912), (1945\space\textbar\space 1939), (1956\space\textbar\space 1961), (1935\space\textbar\space 1923), (1955\space\textbar\space 1949) \\
\cline{2-3}
                            & \small ($en$\space\textbar\space $de^T$) & \footnotesize (1952\space\textbar\space 1949), (1978\space\textbar\space 1977), (1932\space\textbar\space 1941), (1955\space\textbar\space 1953), (1903\space\textbar\space 1922), (1943\space\textbar\space 1956), (1940\space\textbar\space 1939), (1956\space\textbar\space 1961), (1935\space\textbar\space 1923), (1911\space\textbar\space 1901), (1965\space\textbar\space 1962) \\
\cline{2-3}
                            & \small ($en$\space\textbar\space $it^T$) & \footnotesize (1992\space\textbar\space 1985), (1929\space\textbar\space 1932), (1903\space\textbar\space 1921), (1945\space\textbar\space 1935), (1940\space\textbar\space 1943), (1956\space\textbar\space 1961), (1935\space\textbar\space 1923), (1965\space\textbar\space 1962), (1982\space\textbar\space 1986), (1975\space\textbar\space 1969)\\
\cline{2-3}
                            & \small ($en$\space\textbar\space $nl^T$) & \footnotesize (1978\space\textbar\space 1977), (1940\space\textbar\space 1910), (1992\space\textbar\space 1987), (1903\space\textbar\space 1922), (1945\space\textbar\space 1950), (1941\space\textbar\space 1935), (1940\space\textbar\space Het juiste jaartal van de geboorte van schrijver Jeannette Howard Foster is niet beschikbaar.), (1980\space\textbar\space Ik heb geen informatie over een schrijver genaamd Eric San Juan.), (1956\space\textbar\space 1961), (1935\space\textbar\space 1923) \\
\cline{2-3}
                            & \small ($en$\space\textbar\space $sv^T$) & \footnotesize (1943\space\textbar\space 1938), (1941\space\textbar\space 1932), (1952\space\textbar\space 1949), (1978\space\textbar\space 1976), (1932\space\textbar\space Bob McGrath föddes år 1932.), (1992\space\textbar\space Jag beklagar, men jag har ingen information om författaren Zach Hughes och när han föddes.), (1903\space\textbar\space 1921), (1890\space\textbar\space 1871), (1963\space\textbar\space Jag beklagar, men jag kan inte hitta information om författaren Susan Wrights födelseår.), (1943\space\textbar\space 1956) \\
\midrule
\multirow{2}{1.9cm}{{\small \textsc{companies}}} & \small ($en$\space\textbar\space $en$)   & \footnotesize (Dear user, the headquarters of Ford Motor Company is located in Dearborn.\space\textbar\space Dear user, Ford Motor Company has its headquarters in Dearborn.), (Mayfield Village\space\textbar\space Cleveland), (Berlin\space\textbar\space Munich), (Frankfurt\space\textbar\space Mannheim), (Milan\space\textbar\space Genoa), (Rome\space\textbar\space Milan), (Rome\space\textbar\space Milan), (Hilversum\space\textbar\space Amsterdam), (Hoofddorp\space\textbar\space Hague), (Apeldoorn\space\textbar\space Heerenveen) \\
\cline{2-3}
                            & \small ($en$\space\textbar\space $en^P$) & \footnotesize (Issaquah\space\textbar\space Seattle), (North Chicago\space\textbar\space Chicago), (Fort Worth\space\textbar\space Dallas), (Cologne\space\textbar\space Frankfurt), (Stuttgart\space\textbar\space Munich), (Salzgitter\space\textbar\space Salzgitter AG is headquartered in Salzgitter.), (Jena\space\textbar\space Oberkochen), (Nuremberg\space\textbar\space Frankfurt), (San Donato Milanese\space\textbar\space Milan), (Bergamo\space\textbar\space Stezzano) \\
\cline{2-3}
                            & \small ($en$\space\textbar\space $de^T$) & \footnotesize (Issaquah\space\textbar\space Seattle), (Dear user, the headquarters of Ford Motor Company is located in Dearborn.\space\textbar\space Dear user, Ford Motor Company has its headquarters in Dearborn.), (Purchase\space\textbar\space New York City), (Kenilworth\space\textbar\space New York), (Mayfield Village\space\textbar\space Cleveland), (Fort Worth\space\textbar\space Dallas), (Cologne\space\textbar\space Frankfurt), (Selm\space\textbar\space Lünen), (Munich\space\textbar\space Ehningen), (Stuttgart\space\textbar\space Ismaning)  \\
\cline{2-3}
                            & \small ($en$\space\textbar\space $it^T$) & \footnotesize (Irving\space\textbar\space Houston), (Chesterbrook\space\textbar\space Philadelphia), (Issaquah\space\textbar\space Seattle), (Dublin\space\textbar\space Dublino), (Bloomfield\space\textbar\space Philadelphia), (Dear user, the headquarters of Ford Motor Company is located in Dearborn.\space\textbar\space Detroit), (Purchase\space\textbar\space New York City.), (White Plains\space\textbar\space New York.), (Chicago\space\textbar\space Seattle), (North Chicago\space\textbar\space Chicago.) \\
\cline{2-3}
                            & \small ($en$\space\textbar\space $nl^T$) & \footnotesize (Minnetonka\space\textbar\space Minneapolis), (Issaquah\space\textbar\space Seattle), (Dear user, the headquarters of Ford Motor Company is located in Dearborn.\space\textbar\space Dear user, Ford Motor Company heeft zijn hoofdkantoor in Dearborn.), (Deerfield\space\textbar\space De stad waar Walgreens Boots Alliance Inc. zijn hoofdkantoor heeft, is Deerfield.), (Purchase\space\textbar\space New York City), (White Plains\space\textbar\space New York City), (Mayfield Village\space\textbar\space Cleveland), (Fort Worth\space\textbar\space Dallas), (Cologne\space\textbar\space Frankfurt), (Selm\space\textbar\space Lünen) \\
\cline{2-3}
                            & \small ($en$\space\textbar\space $sv^T$) & \footnotesize (Minnetonka\space\textbar\space Minneapolis.), (Dear user, the headquarters of Ford Motor Company is located in Dearborn.\space\textbar\space Dear user, Ford Motor Company's headquarters is located in Dearborn.), (Chicago\space\textbar\space San Francisco.), (Purchase\space\textbar\space New York.), (White Plains\space\textbar\space Hamilton), (Austin\space\textbar\space San Francisco.), (Fremont\space\textbar\space Toronto), (North Chicago\space\textbar\space Chicago), (Stamford\space\textbar\space St. Louis), (Mayfield Village\space\textbar\space Cleveland.) \\
\hline
\end{longtable}

\section{Accuracy scores for ablation experiments}\label{app:F_accuracy_ablation}

\begin{table}[H]
\caption{Accuracy scores for the ablation experiments. We consider ablations in which we translate/paraphrase only the instruction (columns
$I$) or only the input data (columns $X$).}\label{tab:accuracy-ablation}
\centering
\begin{tabular}{l|c|c|c|c|c|c|c|c}
\toprule
{}          &   \multicolumn{2}{c|}{paws} & \multicolumn{2}{c|}{xnli} & \multicolumn{2}{c|}{copa} & \multicolumn{2}{c}{belebele} \\
\midrule
{}          & $I$ & $X$ & $I$ & $X$ & $I$ & $X$ & $I$ & $X$ \\
\midrule
en          & \multicolumn{2}{c|}{75.6} & \multicolumn{2}{c|}{43.7} & \multicolumn{2}{c|}{84.4} & \multicolumn{2}{c}{85.9} \\
\midrule
en$^P$      & 71.8  & 69.7  & 54.2  & 44.5 & 88.2 & 80.6 & 87.4 & - \\
de$^T$      & 62.5  & 78.7  & 49.2  & 38.0 & 88.0 & 82.6 & 86.2 & 80.0 \\
it$^T$      & 72.8  & 76.8  & 57.2  & 38.4 & 91.0 & 81.8 & 86.0 & 78.0 \\
nl$^T$      & 68.7  & 78.9  & 48.1  & 36.8 & 86.4 & 84.8 & 85.0 & 79.2 \\
sv$^T$      & 59.1  & 76.1  & 48.3  & 37.1 & 93.0 & 81.4 & 85.6 & 79.8 \\
\bottomrule
\end{tabular}
\end{table}

\section{Genbench evaluation card}\label{app:A_genbench}

Our work uses generalisation across senses to assess task understanding in LLMs.
In \cref{fig:genbench_eval_card}, we provide the GenBench eval card~\citep{hupkes2023stateoftheart} of our experiments.

\begin{figure}[H]
    \centering
\newcommand{\tabularwidth}{\columnwidth}

\newcommand{\expone}{$\boxtimes$}

\renewcommand{\arraystretch}{1.1}
\setlength{\tabcolsep}{0mm}
\begin{tabular}{|p{\tabularwidth}<{\centering}|}
\hline

\rowcolor{gray!60}
\textbf{Motivation} \\
\footnotesize
\begin{tabular}{p{0.25\tabularwidth}<{\centering} p{0.25\tabularwidth}<{\centering} p{0.25\tabularwidth}<{\centering} p{0.25\tabularwidth}<{\centering}}
\textit{Practical} & \textit{Cognitive} & \textit{Intrinsic} & \textit{Fairness}\\
& 		
& \expone 
& 		

\end{tabular}\\

\rowcolor{gray!60}
\textbf{Generalisation type} \\
\footnotesize
\begin{tabular}{p{0.17\tabularwidth}<{\centering} p{0.20\tabularwidth}<{\centering} p{0.14\tabularwidth}<{\centering} p{0.17\tabularwidth}<{\centering} p{0.18\tabularwidth}<{\centering} p{0.14\tabularwidth}<{\centering}}
\textit{Compositional} & \textit{Structural} & \textit{Cross Task} & \textit{Cross Language} & \textit{Cross Domain} & \textit{Robust- ness}\\
& 		
& 		
& \expone		
& 		
& \hspace{0.2mm} \expone		
\end{tabular}\\

\rowcolor{gray!60}
\textbf{Shift type} \\
\footnotesize
\begin{tabular}{p{0.25\tabularwidth}<{\centering} p{0.25\tabularwidth}<{\centering} p{0.25\tabularwidth}<{\centering} p{0.25\tabularwidth}<{\centering}}
\textit{Covariate} & \textit{Label} & \textit{Full} & \textit{Assumed}\\
\expone\hspace{0.8mm}		
& 		
& 		
& 		

\vspace{2mm} \\
\end{tabular}\\

\rowcolor{gray!60}
\textbf{Shift source} \\
\footnotesize
\begin{tabular}{p{0.25\tabularwidth}<{\centering} p{0.25\tabularwidth}<{\centering} p{0.25\tabularwidth}<{\centering} p{0.25\tabularwidth}<{\centering}}
\textit{Naturally occuring} & \textit{Partitioned natural} & \textit{Generated shift} & \textit{Fully generated}\\
\expone\hspace{0.8mm}		
& 		
& 		
& 		
\vspace{2mm} \\
\end{tabular}\\

\rowcolor{gray!60}
\textbf{Shift locus}\\
\footnotesize
\begin{tabular}{m{0.22\tabularwidth}<{\centering} p{0.28\tabularwidth}<{\centering} p{0.24\tabularwidth}<{\centering} m{0.24\tabularwidth}<{\centering}}
\textit{Train--test} & \textit{Finetune train--test} & \textit{Pretrain-- train} & \textit{Pretrain-- test}\\
& 		
& 		
& \hspace{3.3mm}\expone		
\end{tabular}\\
\hline
\end{tabular}
    \caption{Our experiments assess cross-lingual generalisation for natural corpora, in pretrained LLMs, to assess LLM task understanding.}
    \label{fig:genbench_eval_card}
\end{figure}

\end{appendices}


\end{document}